\pdfoutput=1
\documentclass[utf8]{FrontiersinHarvard} 

\usepackage{url,hyperref,microtype}
\usepackage[onehalfspacing]{setspace}
\usepackage{booktabs}
\usepackage{multirow}
\usepackage{tabularx}
\usepackage{siunitx}
\usepackage{colortbl}      
\usepackage{float}
\usepackage{pifont}        
\usepackage{makecell}
\makeatletter
\providecommand\NewStructureName[1]{}
\providecommand\AssignStructureRole[2]{}
\providecommand\NewTaggingSocket[2]{}
\providecommand\NewTaggingSocketPlug[3]{}
\providecommand\AssignTaggingSocketPlug[2]{}
\providecommand\UseTaggingSocket[1]{}
\providecommand\UseStructureName[1]{}
\providecommand\tagstructbegin[1]{}

\makeatother

\usepackage[most]{tcolorbox}
\usepackage[capitalize]{cleveref}

\definecolor{best}{RGB}{181,229,80}    
\definecolor{second}{RGB}{75,139,190}  
\definecolor{third}{RGB}{236,236,162}  
\newcommand{\best}[1]{\cellcolor{best}\textbf{#1}}
\newcommand{\second}[1]{\cellcolor{second}{#1}}
\newcommand{\third}[1]{\cellcolor{third}{#1}}

\definecolor{promptblue}{RGB}{0,51,102}
\definecolor{promptteal}{RGB}{0,139,139}
\definecolor{promptred}{RGB}{210,40,40}
\definecolor{promptcodebg}{RGB}{235,240,245}
\newcommand{\pcritical}[1]{\textbf{\textcolor{promptred}{#1}}}
\newcommand{\pcode}[1]{\colorbox{promptcodebg}{\texttt{\small\textcolor{promptblue}{#1}}}}
\newtcolorbox{promptcard}[1][]{
  breakable, colback=white, colframe=promptblue, boxrule=0.8pt, arc=2mm,
  left=3mm,right=3mm,top=2mm,bottom=2mm,
  title=#1, coltitle=white, colbacktitle=promptblue, fonttitle=\bfseries
}
\newtcolorbox{promptbox}[1][]{
  colback=white, colframe=promptblue, boxrule=0.8pt, arc=2mm,
  left=3mm,right=3mm,top=2mm,bottom=2mm,
  title=#1, coltitle=white, colbacktitle=promptblue, fonttitle=\bfseries
}

\newcommand{\gf}[1]{}
\newcommand{\Arti}[1]{}
\newcommand{\ma}[1]{}

\def\keyFont{\fontsize{8}{11}\helveticabold }
\def\firstAuthorLast{Al Mdfaa {et~al.}}
\def\Authors{Mohamad Al Mdfaa\,$^{1,*}$, Svetlana Lukina\,$^{1}$, Timur Akhtyamov\,$^{1}$, Arthur Nigmatzyanov\,$^{1}$, Dmitrii Nalberskii\,$^{1}$, Sergey Zagoruyko\,$^{2}$ and Gonzalo Ferrer\,$^{1}$}
\def\Address{$^{1}$Skolkovo Institute of Science and Technology, Moscow, Russia \\
$^{2}$Independent Researcher}
\def\corrAuthor{Mohamad Al Mdfaa}
\def\corrEmail{mohamad.almdfaa@skoltech.ru}

\makeatletter
\def\@maketitle{%
  \let\footnote\thanks
  \clearemptydoublepage
  \checkoddpage\ifcpoddpage\setlength{\aboveskipchk}{-7pc}\else\setlength{\aboveskipchk}{-3pc}\fi%
  \vspace*{\aboveskipchk}%
  \vspace{\dropfromtop}%
  \rule{\textwidth}{1\p@}\par%
  \helvetica
  \hbox to \textwidth{%
  \parbox[t]{36.5pc}{%
    \vspace*{1sp}
    {\helveticabold\fontsize{20}{21}\color{black}\selectfont\raggedright \@title \par}%
    \vspace{4.5\p@}
    {\helveticabold\fontsize{12}{15}\selectfont\raggedright \@author \par}%
    \vspace{4\p@}
    {\helvetica\fontsize{12}{12}\selectfont\raggedright\slshape\@address \Address \\ \par}%
    \vspace{6\p@}
    {\helvetica\fontsize{12}{10}\selectfont\raggedright {Correspondence*:\\ }\@correspondance \corrAuthor \\  \corrEmail \par}
    \vspace{4\p@}
    {\helvetica\fontsize{12}{12}\selectfont\raggedright\@extraAuth \par}%
    \vspace{8\p@}
    }%
  }
  \vspace{14.5\p@}%
}
\def\ps@headings{%
  \def\@oddfoot{\vbox to 12.5\p@{\rule{\textwidth}{0.5\p@}\vss
        \hbox to \textwidth{\helveticabold\small \hfill \thepage}}}%
  \def\@evenfoot{\vbox to 12.5\p@{\rule{\textwidth}{0.5\p@}\vss
        \hbox to \textwidth{\helveticabold\small \hfill \thepage}}}%
  \def\@evenhead{\vbox{\hbox to \textwidth{\fontsize{11}{10}\selectfont
        \helveticabold{\fontshape{it}\selectfont
        \strut\firstAuthorLast}\hfill\strut\rightmark}\vspace{6.5\p@}\rule{\textwidth}{0.5\p@}}}%
  \def\@oddhead{\vbox{\hbox to \textwidth{\fontsize{11}{10}\selectfont
        \helveticabold{\fontshape{it}\selectfont
        \strut\firstAuthorLast}\hfill\strut\rightmark}\vspace{6.5\p@}\rule{\textwidth}{0.5\p@}}}%
  \def\titlemark##1{\markboth{##1}{##1}}%
  \def\authormark##1{\gdef\leftmark{##1}}%
}
\ps@headings
\makeatother

\begin{document}
\onecolumn
\firstpage{1}

\title[Persistent Spatiotemporal Knowledge Graphs]{Spatiotemporal Knowledge Graphs as Persistent Scene Memory for Embodied Question Answering}

\author[\firstAuthorLast ]{\Authors}
\address{}
\correspondance{}
\extraAuth{}

\maketitle

\begin{abstract}
\section{}
Vision-language models (VLMs) demonstrate strong image-level scene understanding, but reasoning over long egocentric video remains costly: because VLMs maintain no persistent memory or explicit spatial representation, all sampled frames must be re-processed for every new query. We present VL-KnG, a training-free framework that constructs \textit{spatiotemporal knowledge graphs} from monocular egocentric video, bridging fine-grained scene graphs and global topological graphs without 3D reconstruction. VL-KnG processes video in chunks, maintains persistent object identities via large language model (LLM)-based Spatiotemporal Object Association (STOA), and answers questions through Graph-Enhanced Retrieval (GER), which combines subgraph retrieval with visual grounding. Once constructed, the knowledge graph removes the need to revisit raw video at query time, decoupling query latency from video length. Across three embodied question answering benchmarks---OpenEQA, NaVQA, and our newly introduced \textit{WalkieKnowledge}---VL-KnG achieves accuracy competitive with frontier VLMs while answering queries at substantially lower latency and with explainable, graph-grounded reasoning; it further surpasses prior persistent-representation baselines and open-weight VLMs in several settings. Deployment on a real robot demonstrates practical applicability, with query latency remaining stable as observation history grows. As a persistent, queryable scene memory, VL-KnG provides a concrete substrate for memory maintenance and knowledge updating in embodied AI agents.

\tiny
\keyFont{\section{Keywords:} spatiotemporal knowledge graph, vision-language models, retrieval-augmented generation, video question answering, persistent memory, embodied agents, robot navigation, egocentric video}
\end{abstract}

\section{Introduction}
\label{sec:intro}
\textit{Embodied scene understanding from egocentric video} is the problem of building an internal representation of an environment from first-person visual observations, such that an embodied agent can later reason about---and answer natural-language questions about---the objects, spatial layout, and events it has observed~\citep{das2018embodied, majumdar2024openeqa}. We address its \textit{episodic-memory} form~\citep{majumdar2024openeqa}, in which the agent observes the environment once and cannot revisit it at question time, so every answer must be grounded in what was retained from that traversal. For example, consider a robot that has walked through a building and is later asked, ``where did you last see the fire extinguisher?'' or ``was there a chair near the entrance?''. Answering requires remembering what was observed, where objects were located, and how the scene was organized---often long after the moment of observation.

Vision-language models (VLMs)~\citep{kaduri2025s, liu2025survey, comanici2025gemini, bai2025qwen2} excel at understanding individual images, yet they are poorly suited to this setting. They maintain no persistent, structured memory of a scene; their spatial reasoning remains implicit and difficult to explain; and, because they must re-process the sampled frames for every new question, their cost grows as $O(\text{queries} \times \text{frames})$ and scales poorly with video length.

\begin{figure}[htbp]
    \centering
    \includegraphics[width=0.9\linewidth]{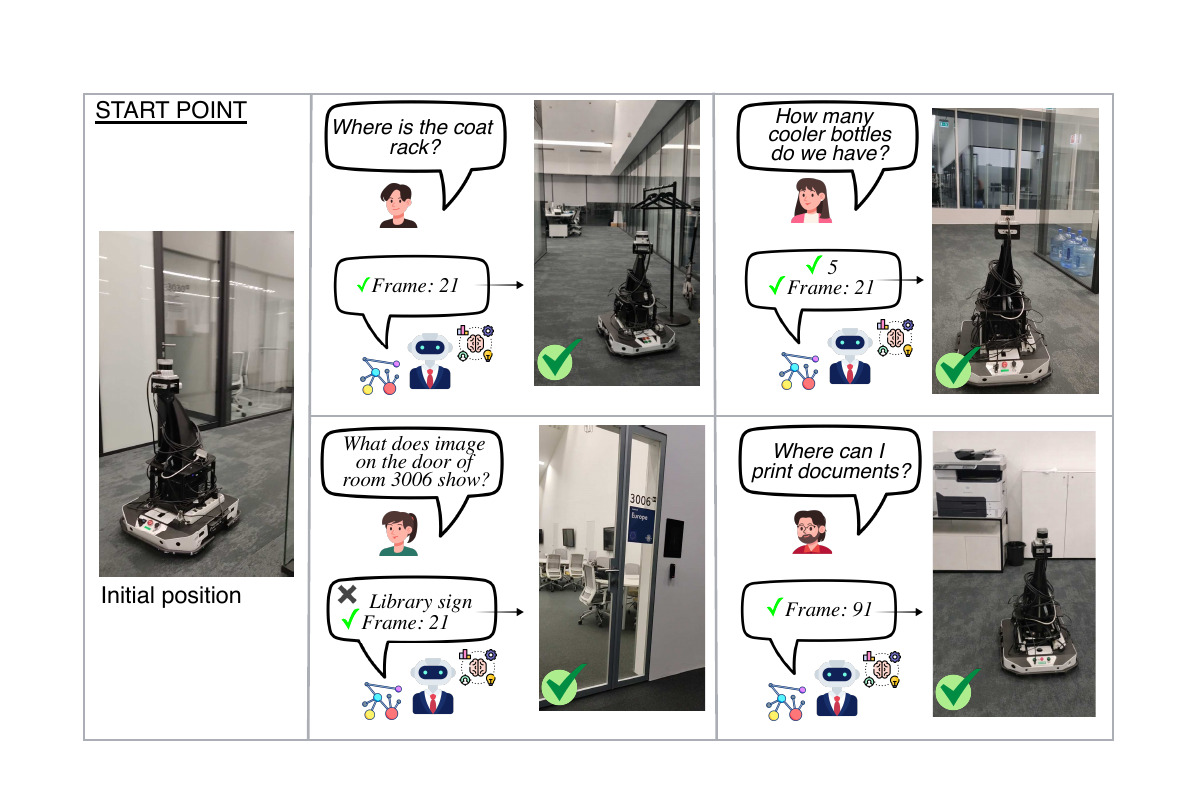}
    \caption{Real-world deployment examples of VL-KnG for embodied scene understanding. The system processes natural language queries over a spatiotemporal knowledge graph to identify relevant objects and provide frame-level localization, enabling downstream tasks such as navigation goal identification.}
    \label{fig:real_world_grid}
\end{figure}

We introduce \textbf{VL-KnG} (Vision-Language Knowledge Graph)\footnote{Code and benchmark will be released upon acceptance.}, a training-free system that processes an egocentric video \emph{once} and converts it into a persistent, queryable representation of the scene, namely a \textit{spatiotemporal knowledge graph} whose nodes are the individual objects observed by the camera and whose edges capture their spatial relationships over time. VL-KnG performs three functions. (i)~It \textit{constructs} the graph from ordinary monocular RGB video---without depth sensing, 3D reconstruction, or task-specific training---by chunking the video and detecting objects with a VLM. (ii)~It \textit{maintains consistent object identity} even when an object reappears much later in a visually unrelated part of the trajectory, using an LLM to decide which detections correspond to the same object instance (\textit{Spatiotemporal Object Association}, STOA). (iii)~It \textit{answers questions} by retrieving the scene subgraphs of the most relevant frames via lightweight lexical matching over object descriptions and passing them to an LLM for reasoning---graph-based retrieval-augmented generation (GraphRAG)~\citep{Han2024RetrievalAugmentedGW}---optionally augmented with SigLIP2-based~\citep{tschannen2025siglip} visual grounding for improved frame localization (\textit{Graph-Enhanced Retrieval}, GER). As shown in \Cref{fig:real_world_grid}, this enables a real robot to answer language queries over the graph and localize the goal object directly in the video stream.

The key benefit is that the graph is constructed only once, at cost $O(\text{frames})$, after which each query is answered by lightweight retrieval over the graph rather than by re-ingesting video frames---decoupling per-query cost from video length. The retrieved subgraph also serves as an inspectable reasoning trace, in contrast to the opaque end-to-end reasoning of VLMs, giving explicit structured representations complementary advantages in explainability~\citep{tiddi2022knowledge}, efficiency, and adaptability across downstream tasks.

From a representation perspective, prior approaches typically fall into two categories. 
Fine-grained scene graphs~\citep{3dgraphllm2025iccv, fross2025iccv} capture detailed object-level semantics but are usually local and often rely on 3D reconstruction, while topological graphs~\citep{chiang2024mobility, gsavln2025iclr} model large environments but lack rich object-level semantics. 
\textit{VL-KnG unifies these paradigms by constructing spatiotemporal knowledge graphs that preserve object-level detail and rich spatial relationships while scaling to large environments from monocular video}, across the entire trajectory.

Concretely, graph construction runs modern VLMs~\citep{comanici2025gemini, bai2025qwen2} over video chunks and STOA links detections across time into persistent objects with rich semantic and spatial relationships, while query processing retrieves the scene subgraphs of the most relevant frames via lightweight lexical matching over object descriptions and passes them to an LLM for reasoning. GER further augments this retrieval with SigLIP2-based~\citep{tschannen2025siglip} visual grounding for improved frame localization.

We evaluate VL-KnG on three benchmarks spanning diverse environments:
\textbf{OpenEQA}~\citep{majumdar2024openeqa} (1,636 embodied QA pairs across 180+ indoor environments),
\textbf{NaVQA}~\citep{anwar2024remembr} (descriptive QA over indoor/outdoor robot navigation sequences), and
\textbf{WalkieKnowledge}, our new proposed benchmark with 193 manually annotated questions across 8 egocentric trajectories.
Our contributions are:
\begin{itemize}
\item \textbf{A training-free pipeline for constructing spatiotemporal knowledge graphs from monocular video}, using VLM-based detection~\citep{comanici2025gemini} and LLM-based Spatiotemporal Object Association (STOA) to maintain persistent object identity---without 3D reconstruction, depth sensing, or task-specific training.
\item \textbf{Graph-Enhanced Retrieval (GER)}---a hybrid approach combining Graph\-RAG~\citep{Han2024RetrievalAugmentedGW} subgraph retrieval with SigLIP2~\citep{tschannen2025siglip} visual grounding for improved frame localization.
\item \textbf{WalkieKnowledge benchmark} with 193 manually annotated questions across 8 diverse egocentric trajectories enabling comparison between structured approaches and general-purpose VLMs.
\end{itemize}

\section{Related Work}
\label{sec:related}
\subsection{Spatiotemporal Scene Graphs from Video}
Scene graphs encode objects as nodes and relationships as edges.
Recent work also extends scene graphs temporally. SceneSayer~\citep{scenesayer2024eccv} predicts scene graph evolution, DriveLM~\citep{drivelm2024eccv} introduces graph-structured VQA for driving, and STEP~\citep{step2025cvpr} constructs spatio-temporal scene graphs for Video-LLMs.
In 3D, Lost~\&~Found~\citep{lostfound2025cvpr} builds dynamic 3D scene graphs from egocentric observations, FROSS~\citep{fross2025iccv} enables real-time 3D scene graph generation from monocular video, and 3DGraphLLM~\citep{3dgraphllm2025iccv} combines 3D scene graphs with LLMs for spatial understanding.
Our \textit{spatiotemporal knowledge graphs} combine the strengths of both families. From scene graphs, we inherit fine-grained object-level detail with rich semantic descriptors and spatial relationships, while from topological graphs, we inherit the ability to cover large environments from monocular video without 3D reconstruction. By adding persistent object identity via LLM-based semantic association, our representation bridges these two paradigms in a way not explored in prior work.

\subsection{Persistent Memory from Egocentric Video}
Maintaining long-horizon memory from egocentric observations is critical for embodied agents.
VideoAgent~\citep{videoagent2024eccv} iteratively searches long videos using structured memory, AMEGO~\citep{amego2024eccv} constructs active memory representations, and Embodied VideoAgent~\citep{embodiedvideoagent2025iccv} integrates persistent scene memory with embodied sensor streams.
ReMEmbR~\citep{anwar2024remembr} builds retrieval-augmented spatio-temporal memory pairing visual observations with metric poses, while SNOW~\citep{snow2025arxiv} constructs 4D scene graphs with persistent object identity.
VL-KnG builds persistent memory through knowledge graphs rather than latent representations, enabling explicit, queryable, and interpretable scene memory that is queried without re-processing the video.

\subsection{Environment Representations for Navigation and Embodied QA}
Vision-language navigation (VLN)~\citep{wu2024vision, anderson2018vision} connects navigation with language instructions, and the choice of environment representation largely determines which queries an agent can answer. Multimodal 3D-mapping methods such as VLMaps~\citep{huang2022visual} and ConceptFusion~\citep{jatavallabhulaconceptfusion} augment metric 3D maps---widely used in robotics---with multimodal embeddings, enabling natural-language queries over the map, and ConceptGraphs~\citep{gu2024conceptgraphs} extends this line to multimodal 3D scene graphs for LLM-based reasoning. 3D scene graphs~\citep{armeni20193d} more broadly underpin systems such as Hydra~\citep{hughes2022hydra} and Clio~\citep{maggio2024clio}, but all of these rely on depth or range sensing; RoboHop~\citep{garg2024robohop} takes a step toward removing this requirement by building a topological graph over image segments.

Image-based topological graphs~\citep{shah2021ving} avoid range sensing, where full images of visited locations serve as nodes, and edges carry traversability scores. Such graphs scale to large environments but lack fine-grained object detail. LM-Nav~\citep{shah2023lm} selects image goals via CLIP-based retrieval and hands them to a learned local policy; NavGPT-2~\citep{zhou2024navgpt} combines LLM reasoning with topological graphs; GSA-VLN~\citep{gsavln2025iclr} maintains persistent topological graph memory; and MobilityVLA~\citep{chiang2024mobility} builds a topological graph from a demonstration tour video, passing the same video to a large VLM to identify a goal frame. World-model approaches predict future observations or states for zero-shot navigation (NWM~\citep{nwm2025cvpr}, WMNav~\citep{nie2025wmnav}), and UniGoal~\citep{unigoal2025cvpr} unifies zero-shot goal navigation via scene-graph representations. On the embodied QA side, OpenEQA~\citep{majumdar2024openeqa} establishes a comprehensive benchmark for scene understanding from egocentric histories, and ReMEmbR~\citep{anwar2024remembr} selects navigation goals via retrieval-augmented memory over visited frames.

VL-KnG inherits complementary strengths from each group, where it constructs a knowledge graph from a single tour video, capturing object-level detail together with environment-scale coverage, and passes retrieved subgraphs to an LLM for question answering and goal-frame proposal, whose output can feed a vision-only policy or a classical navigation stack.

\subsection{Open-Vocabulary Scene Understanding}
Open-vocabulary approaches enable scene understanding without predefined categories.
ConceptGraphs~\citep{gu2024conceptgraphs} constructs open-vocabulary 3D scene graphs using foundation model features, OvSGTR~\citep{ovsgtr2024eccv} addresses open-vocabulary scene graph generation via transformers, and SceneGraphLoc~\citep{scenegraphloc2024eccv} leverages scene graphs for visual localization. UPPM~\citep{robotics15020031} builds persistent, promptable panoptic maps by reconciling open-vocabulary labels with a unified category structure via foundation-model-based dynamic descriptors, fused within a multi-resolution TSDF map without additional training.
Unlike the 3D-based approaches among these, which require depth sensing or 3D reconstruction, VL-KnG operates on monocular RGB video, building open-vocabulary knowledge graphs using VLM-based detection and semantic relationship extraction.

\subsection{Graph-Based Retrieval and Reasoning over Video}
\label{sec:concurrent}

GraphRAG~\citep{Han2024RetrievalAugmentedGW} has emerged as a powerful paradigm for structured retrieval and reasoning, and a growing body of concurrent work couples graph- or retrieval-augmented structure with video understanding. Several systems build their structure at the level of clips, captions, scenes, or transcript-derived text entities rather than persistent visual objects. Vgent~\citep{vgent2025neurips} forms a graph whose nodes are fixed-length video clips linked by text-entity similarity and runs a large vision--language model in a reasoning loop at query time; CLiViS~\citep{clivis2025} maintains a cognitive map that is rebuilt per query inside an iterative LLM--VLM loop; and GraphVideoAgent~\citep{graphvideoagent2025} constructs and mutates a caption-entity graph inside a per-query agent loop. VideoRAG~\citep{videorag2025} indexes a multi-video corpus through ASR-mined cross-video text entities, and SceneRAG~\citep{scenerag2025} segments videos into narrative scenes by prompting an LLM over ASR transcripts; both depend on speech or subtitle signals and are not defined for silent egocentric or robot video.

A second group operates, like VL-KnG, at the level of visual objects. RAVU~\citep{ravu2025} is closest to our representation because it builds a once-per-video spatiotemporal entity graph and answers multi-hop VideoQA by executing a fixed library of compositional reasoning operators over it. Like VL-KnG, RAVU relies on a VLM for open-vocabulary detection, but it maintains entity identity with a SAM~2~\citep{ravi2025sam} visual tracker that links detections into tracklets across temporally contiguous frames, inheriting the visual-continuity assumption and tracker dependence that VL-KnG avoids. Moreover, its nodes are action-centric entities for activity QA rather than an environment map with explicit spatial relations, and it produces neither pose-grounded outputs nor navigation goals. SAMJAM~\citep{samjam2025} is likewise a training-free, monocular pipeline that builds egocentric scene graphs with a VLM, but it propagates object identity with SAM~2 masks within visually continuous clips and targets scene-graph \emph{generation} rather than question answering. DAAAM~\citep{daaam2025arxiv} constructs 4D scene graphs for temporally grounded QA but requires multi-modal sensing beyond monocular RGB. On the egocentric-memory side, Ego-Topo~\citep{egotopo2020} builds a topological graph of activity \emph{zones} for affordance and action anticipation---nodes are zones, not object instances, and it performs no natural-language QA---while OSNOM/LMK~\citep{osnom2025} produces continuous 3D object tracks, including out-of-sight objects, but does not expose a queryable semantic graph. Lastly, VideoAgent~\citep{videoagent2024eccv} is an action-centric long-video QA agent evaluated via clip captions on EgoSchema~\citep{NEURIPS2023_90ce332a}/NExT-QA~\citep{Xiao_2021_CVPR}; because it targets a different task and input regime than persistent object-centric embodied memory, we treat it as related rather than as a directly comparable baseline.

In contrast, VL-KnG,as shown in \Cref{tab:rel}, operates purely on \emph{monocular RGB} video, without depth sensing or 3D reconstruction, yet (i)~builds a \emph{once-built}, object-instance knowledge graph that is queried without re-processing the video; (ii)~maintains persistent object identity across temporally distant, visually non-overlapping chunks through LLM-based semantic association (STOA) rather than visual tracking; and (iii)~answers language QA by single-shot subgraph retrieval over that persistent graph, additionally exposing pose-grounded goal frames (frame index, bounding box, and pose) for downstream robotic use.

\begin{table}[htbp]
\centering\footnotesize
\setlength{\tabcolsep}{3pt}
\renewcommand{\arraystretch}{1.15}
\caption{Mechanism-level comparison of VL-KnG with concurrent graph/retrieval video systems, navigation/memory methods, and scene-graph systems. ``Monocular RGB'' means that the representation is built purely from monocular RGB video, without depth sensing or 3D reconstruction. ``Once-built'' means that the representation is constructed once and queried without re-processing the video. ``Persistent obj.\ ID'' denotes persistent object-instance identity. ``Identity across gaps'' means that identity is maintained across temporally distant observations with no visual overlap, i.e., without frame-to-frame tracking or mask propagation.}
\label{tab:rel}
\begin{tabular}{@{}llccccc@{}}
\toprule
Work & Graph / node type & \makecell{Monocular\\RGB} & \makecell{Once-\\built} & \makecell{Persistent\\obj.\ ID} & \makecell{Language\\QA} & \makecell{Identity across\\gaps}\\
\midrule
Vgent~\citep{vgent2025neurips} & video clips & \ding{51} & \ding{55} & \ding{55} & \ding{51} & \ding{55}\\
CLiViS~\citep{clivis2025} & text entities & \ding{51} & \ding{55} & \ding{55} & \ding{51} & \ding{55}\\
Ego-Topo~\citep{egotopo2020} & activity zones & \ding{51} & \ding{51} & \ding{55} & \ding{55} & \ding{55}\\
RoboHop~\citep{garg2024robohop} & topological (segments) & \ding{51} & \ding{51} & \ding{55} & \ding{55} & \ding{55}\\
WMNav~\citep{nie2025wmnav} & value map (online) & \ding{55} & \ding{55} & \ding{55} & \ding{55} & \ding{55}\\
ReMEmbR~\citep{anwar2024remembr} & frame memory (+poses) & \ding{51} & \ding{51} & \ding{55} & \ding{51} & \ding{55}\\
RAVU~\citep{ravu2025} & entity spatio-temporal graph & \ding{51} & \ding{51} & \ding{51} & \ding{51} & \ding{55}\\
ConceptGraphs~\citep{gu2024conceptgraphs} & 3D object scene graph & \ding{55} & \ding{51} & \ding{51} & \ding{51} & \ding{51}\\
DAAAM~\citep{daaam2025arxiv} & 4D scene graph & \ding{55} & \ding{51} & \ding{51} & \ding{51} & \ding{51}\\
SAMJAM~\citep{samjam2025} & per-frame SAM2 masks & \ding{51} & \ding{51} & \ding{51} & \ding{55} & \ding{55}\\
OSNOM/LMK~\citep{osnom2025} & 3D object tracks (no graph) & \ding{55} & \ding{51} & \ding{51} & \ding{55} & \ding{51}\\
\midrule
\textbf{VL-KnG (ours)} & \textbf{object graph + spatial relations} & \textbf{\ding{51}} & \textbf{\ding{51}} & \textbf{\ding{51}} & \textbf{\ding{51}} & \textbf{\ding{51}}\\
\bottomrule
\end{tabular}
\end{table}

\section{Problem Formulation}
\label{sec:problem}
We consider embodied scene understanding from egocentric video in its episodic-memory form in which the environment is observed once and questions are asked afterward, requiring answers to rely on information retained during the traversal.
The input consists of an egocentric video, recorded by a robot or a human, given as a sequence of frames $\mathcal{I} = \{I_t\}_{t=1}^{T}$ with $I_t \in \mathbb{R}^{H \times W \times 3}$, and a set of natural language queries $\mathcal{Q} = \{q_n\}_{n=1}^{N}$ posed after the video has been observed.
For each query $q_n$, the system must produce an answer $a_n$ and the frame indices $\mathcal{F}_n \subseteq \{1, \ldots, T\}$ most relevant to the query, i.e., the frames containing the objects or scene elements the query refers to.

We approach this problem by constructing a spatiotemporal knowledge graph $\mathcal{G} = (V, E)$ that represents the observed environment.
Each node $o_i \in V$ is a unique physical object, characterized by a structured descriptor
\[
o_i = \left(\textit{id}_i,\ \mathcal{F}_i,\ \{\textit{bbox}_{i,t}\}_{t \in \mathcal{F}_i},\ \textit{color}_i,\ \textit{material}_i,\ \textit{size}_i,\ \textit{affordances}_i\right),
\]
where $\mathcal{F}_i \subseteq \{1, \ldots, T\}$ is the set of frames in which the object was observed and $\textit{bbox}_{i,t}$ is its bounding box in frame $t$.
Each edge $e = (o_i, r, o_j, \mathcal{F}_e) \in E$ is a typed spatial relation between two objects, where $\mathcal{F}_e \subseteq \mathcal{F}_i \cap \mathcal{F}_j$ records the frames in which the relation was observed, and $r$ is drawn from the following set of 20 spatial predicates.
\begin{multline*}
  \{\textit{on},\ \textit{on top of},\ \textit{under},\ \textit{next to},\ \textit{between},\ \textit{in front of},\ \textit{behind},\ \textit{near},\ \textit{far from},\ \textit{touching},\\
  \textit{separate from},\ \textit{left of},\ \textit{right of},\ \textit{above},\ \textit{below},\ \textit{inside},\ \textit{outside},\ \textit{surrounding},\ \textit{adjacent to},\ \textit{against}\}
\end{multline*}
Relations are extracted per frame during graph construction and accumulated across chunks; predicates involving more than two objects, such as \textit{between}, are stored as pairwise instances against each reference object.
Thus, our objective is to develop: (i)~a procedure that builds $\mathcal{G}$ from a given video $\mathcal{I}$ in a single pass and (ii)~another procedure that, given $\mathcal{G}$ and a query $q_n$, retrieves the answer $a_n$ and frame indices $\mathcal{F}_n$ without re-processing the video.

The central challenge is maintaining \textit{spatiotemporal consistency}---the same physical object must be assigned a single identity in $\mathcal{G}$ even when it appears across multiple video chunks under varying viewpoints, lighting, and occlusion conditions. This persistent identity is a prerequisite for answering queries that require reasoning over both spatial relationships (``what is next to the couch?'') and temporal information (``when did the car appear?'').
Our approach addresses this through Spatiotemporal Object Association (STOA), which uses LLM-based semantic reasoning to establish object correspondences across chunks, yielding a unified knowledge graph that supports structured retrieval and reasoning.

\section{Method}
\label{sec:method}

\begin{figure}[htbp]
    \centering
    \includegraphics[width=0.9\textwidth]{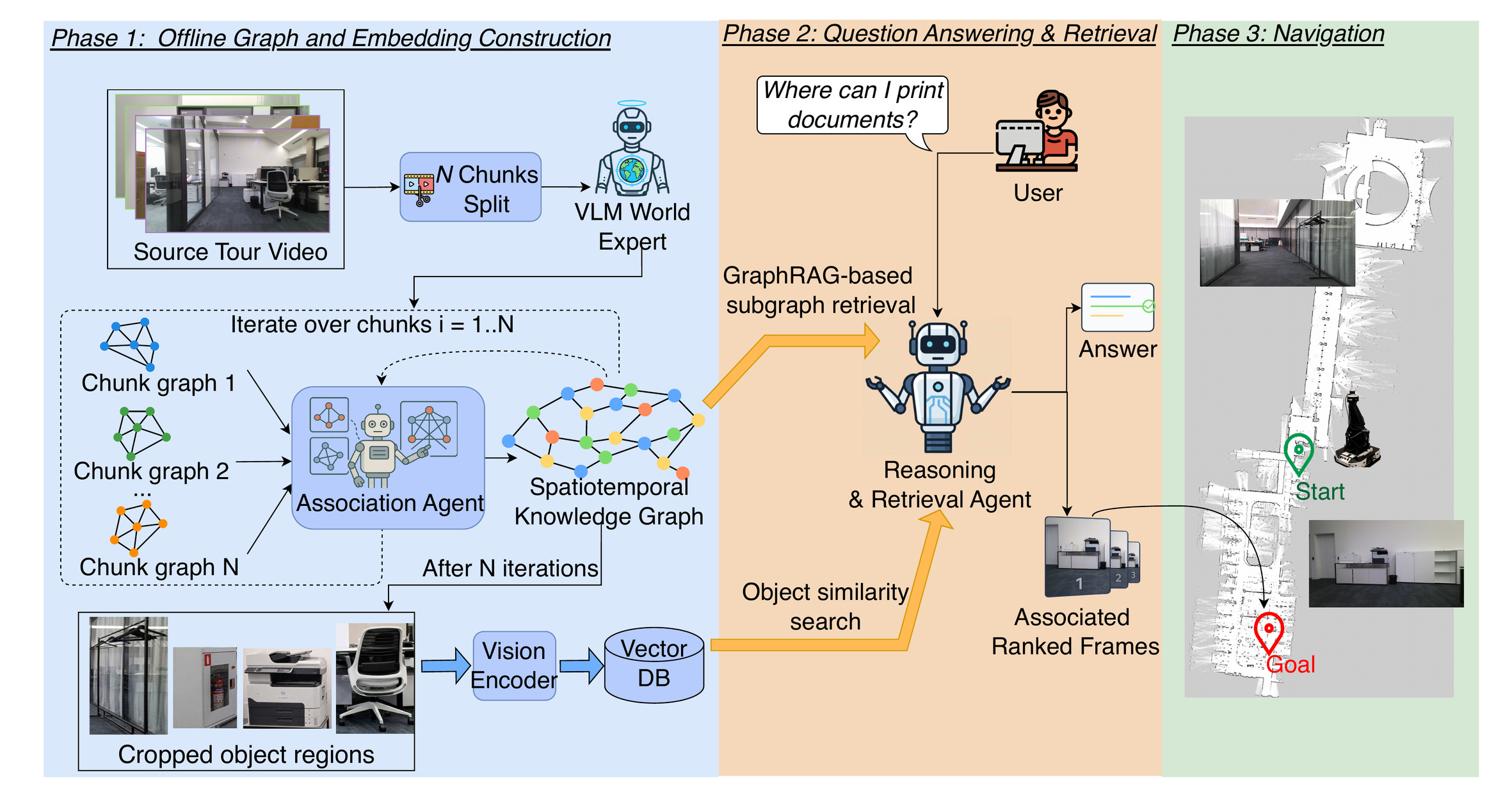}
    \caption{VL-KnG system architecture. In \textbf{Phase~1} (offline), a source tour video is split into frame chunks, each processed by a VLM World Expert to produce chunk graphs; an Association Agent merges these into a unified spatiotemporal knowledge graph, while a Vision Encoder (SigLIP2) embeds cropped object regions into a vector database. In \textbf{Phase~2} (online), given a user query, a Reasoning \& Retrieval Agent combines GraphRAG subgraph retrieval with object similarity search to produce an answer and ranked frames. In \textbf{Phase~3}, the top-ranked goal frame provides a navigation target for downstream robotic use.}
    \label{fig:pipeline}
\end{figure}

VL-KnG operates in two phases (\cref{fig:pipeline}). The offline phase constructs a spatiotemporal knowledge graph and an object embedding database from a source video, while the online phase uses a reasoning agent to answer user queries via GraphRAG subgraph retrieval and visual similarity search, producing both textual answers and ranked goal frames for navigation.

\subsection{Spatiotemporal Knowledge Graph Construction}
The knowledge graph construction process begins with chunking of video frames to maintain temporal consistency while ensuring computational efficiency.
Given a video sequence $\mathcal{I} = \{I_t\}_{t=1}^T$, we partition it into $K = \lceil T/b \rceil$ chunks $\mathcal{C}_k = \{I_{kb+1}, \ldots, I_{\min((k+1)b,\,T)}\}$ for $k = 0, \ldots, K-1$, where $b$ is the chunk size and the last chunk may contain fewer than $b$ frames. We use $b=8$ in all experiments; the effect of this choice is analyzed in \Cref{sec:chunk_size_ablation}.

For each chunk $\mathcal{C}_k$, we employ a vision-language model with multi-image prompting capabilities~\citep{comanici2025gemini, bai2025qwen2} to extract object descriptors $\mathcal{O}_k = \{o_i^k\}_{i=1}^{M_k}$, where $M_k$ is the number of objects detected in chunk $k$.\footnote{The full prompt templates are provided in the Supplementary Material.} These object descriptors form a \textit{chunk graph} $\mathcal{G}_k^{\textit{chunk}}$, which can be considered a `local' knowledge graph covering the frames of chunk $k$ only.
We build the final knowledge graph $\mathcal{G}$ iteratively, processing chunks one by one and denoting the accumulated knowledge graph at iteration $k$ by $\mathcal{G}^{(k)}$. At chunk $k=0$, the chunk graph $\mathcal{G}_0^{\textit{chunk}}$ is obtained, and we initialize $\mathcal{G}^{(0)} \leftarrow \mathcal{G}_0^{\textit{chunk}}$. On subsequent iterations, the graph is updated as
\begin{equation}
\mathcal{G}^{(k)} \leftarrow \text{STOA}(\mathcal{G}^{(k-1)}, \mathcal{G}_k^{\textit{chunk}}),
\end{equation}
where $\text{STOA}$ denotes the spatiotemporal object association procedure described in \cref{sec:soa}. The knowledge graph $\mathcal{G}^{(K-1)}$ serves as the final environment knowledge graph $\mathcal{G}$, stored in a graph database~\citep{miller2013graph} used in the subsequent stages of the pipeline.
This structured representation enables efficient spatial reasoning through graph traversal operations, providing a persistent memory of the environment that is queried without re-processing the video.

\subsection{Spatiotemporal Object Association} \label{sec:soa}
Maintaining object identity over time is crucial for coherent scene understanding.
Traditional approaches rely on visual similarity metrics, which often fail when objects undergo appearance changes due to lighting, occlusion, or viewpoint variations.
Recent geometry-grounded methods such as SegMASt3R~\citep{jayanti2025segmastr} and propagation-based trackers like SAM~2~\citep{ravi2025sam} address segment matching across frames using visual and spatial cues.
However, in our chunk-based pipeline, temporally distant observations may lack visual overlap, motivating a semantic association approach based on reasoning over textual object descriptions.
We propose a semantic-based association mechanism that leverages large language model reasoning~\citep{comanici2025gemini, bai2025qwen2} to establish object correspondences across chunks.

Association is performed at the graph level. Given the accumulated graph $\mathcal{G}^{(k-1)}$ with node set $V^{(k-1)}$ and the new chunk graph $\mathcal{G}_k^{\textit{chunk}}$, the language model jointly decides, for each object in the chunk, whether it matches an existing identity or introduces a new one:
\begin{equation}
\text{Assoc}(o_j^{k} \mid \mathcal{G}^{(k-1)}) \in \{\,\text{match}(o_i) : o_i \in V^{(k-1)}\,\} \cup \{\,\text{new}\,\},
\end{equation}
reasoning over object type, attributes (color, material, size, affordances), and relational context rather than thresholding an explicit similarity score. This makes association robust to appearance changes across viewpoints, lighting, and occlusion, maintaining temporal consistency in the knowledge graph.

\begin{figure}[htbp]
    \centering
    \includegraphics[width=\textwidth]{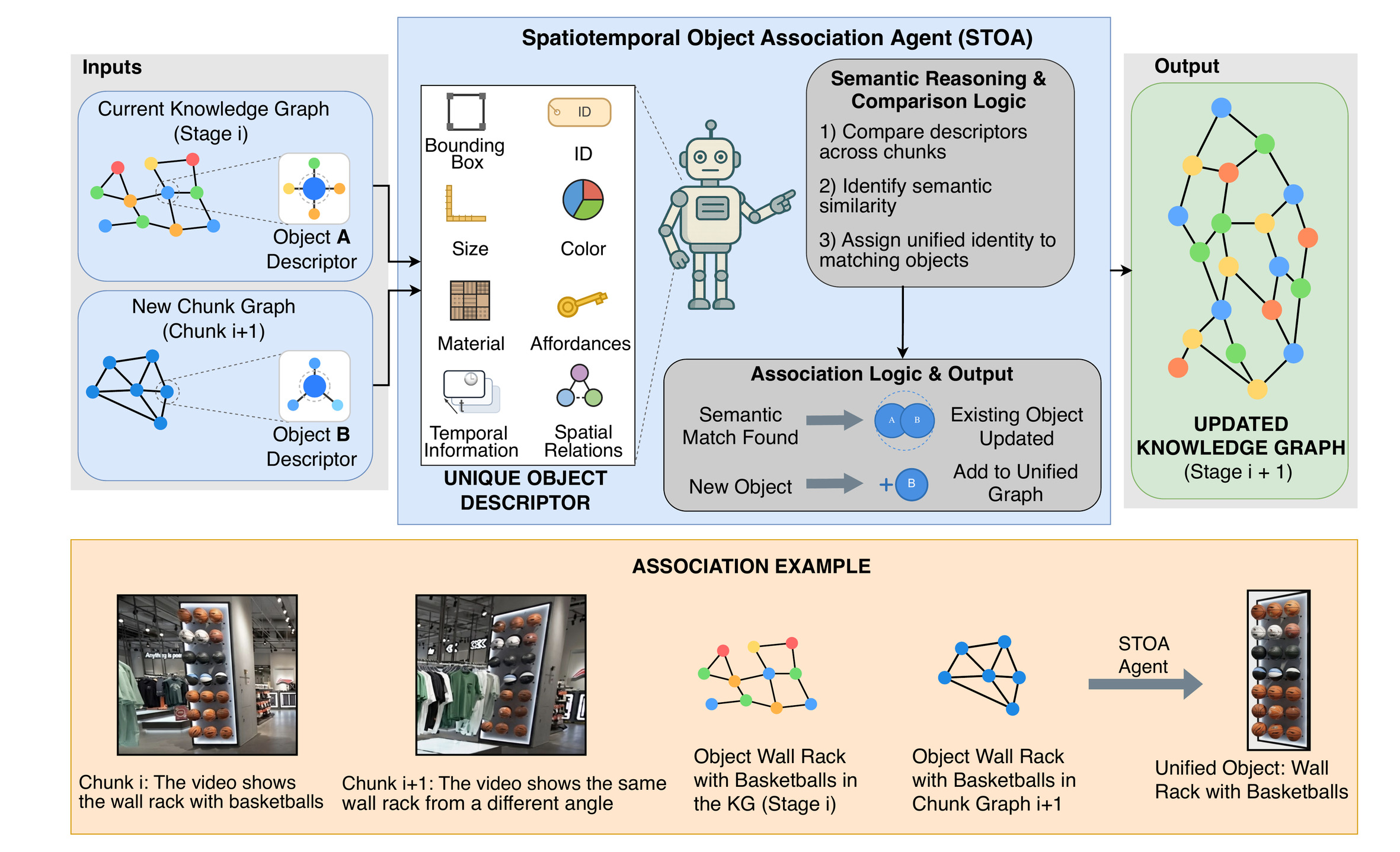}
    \caption{The Spatiotemporal Object Association (STOA) agent. Given the current global knowledge graph and the graph extracted from the incoming video chunk, the LLM-based agent matches detections by type, description, and relational context, reusing existing object IDs for re-observed entities and assigning new IDs to genuinely new objects.}
    \label{fig:stoa_agent}
\end{figure}

In implementation, the association operates on textual object descriptors rather than pixels, realized as the first two steps of our four-step prompt pipeline (Steps~3--4 support query processing, \Cref{sec:query}; the complete templates are given in the Supplementary Material). \emph{Chunk detection} (Step~1) instructs the model to emit, for each chunk, a structured list of objects with a globally incrementing identifier of the form \texttt{<type>\_<num>}, per-frame bounding boxes, and the spatial relations listed in \Cref{sec:problem}. \emph{Cross-chunk resolution} (Step~2) then aligns the new chunk's local identifiers with the existing graph under a simple rule set---``same type with a similar description $\Rightarrow$ reuse the existing global ID; a clearly new object $\Rightarrow$ assign a new ID; when in doubt, prefer reuse''---and returns only the corrected, identifier-aligned descriptors. Because matching operates on appearance, temporal continuity, and relational context rather than absolute position, an object that moves between chunks is linked to its prior identity and simply receives an updated bounding box in the later chunk.
\Cref{fig:stoa_agent} illustrates this process. At each stage, the STOA agent receives the current global knowledge graph together with the incoming chunk graph and decides which detections correspond to already-known physical entities and which should be introduced as new nodes. A step-by-step visualization of graph construction on a single episode is provided in Supplementary Figure~S1.

\subsection{Query Processing} \label{sec:query}

The query processing pipeline follows the graph-based retrieval-augmented generation (GraphRAG) paradigm~\citep{Han2024RetrievalAugmentedGW}, retrieving the local scene subgraphs of the frames most relevant to a query and passing them as structured context to an LLM.
Unlike corpus-scale GraphRAG variants that rely on community detection and hierarchical community summarization, our setting is a compact, vision-built graph, so we adopt a lightweight retrieval scheme that ranks frames by query--object relevance and returns their full local subgraphs.
Given a natural language query $q_n$, the system performs the following steps:
\begin{enumerate}
    \item \textbf{Query--Graph Alignment:} Search terms are extracted from $q_n$ by stop-word filtering, optionally augmented by an LLM step that rewrites the query to reference known knowledge-graph object names, aligning free-form language with the graph vocabulary.
    \item \textbf{Subgraph Retrieval:} Each object is scored by lexical overlap between the query terms and a textual representation of the object (its name, description attributes, and spatial-position phrases). Object scores are accumulated per frame, and frames are ranked by aggregate relevance, with an optional boost for frames containing spatial relations when the query is spatial. The top-$m$ frames are returned as subgraphs $\mathcal{G}_{\textit{sub}} \subseteq \mathcal{G}$, each carrying its complete frame-local context (all objects and their spatial relationships), not only the matched objects.
    \item \textbf{Reasoning and Localization:} The retrieved subgraphs are serialized to text and processed by an LLM to produce the answer $a_n$ and rank the candidate frames $\mathcal{F}_n$ for localization, considering both spatial relationships and temporal ordering.
\end{enumerate}
While this pipeline is demonstrated for embodied QA and navigation goal identification, the graph-based reasoning is general and applicable to other tasks requiring structured scene understanding from video.

\subsection{Graph-Enhanced Retrieval (GER)} \label{sec:ger}
To improve frame localization beyond purely graph-based retrieval, we introduce Graph-Enhanced Retrieval with Object-Level Visual Grounding (GER).
This approach augments the GraphRAG retrieval pipeline with visual similarity search over object-level SigLIP2 embeddings~\citep{tschannen2025siglip}.

Given a query, relevant objects are retrieved through two complementary mechanisms:
\begin{enumerate}
    \item[(i)] \textbf{Graph-based retrieval} over the knowledge graph, as described above, yielding a set of semantically matched objects and their associated frames.
    \item[(ii)] \textbf{Visual similarity search} over SigLIP2 embeddings extracted from detected object bounding boxes. For each detected object in the knowledge graph, we extract its bounding box region from the corresponding video frame and compute an image embedding. At query time, the query text is encoded with the same vision-language encoder, and cosine similarity is computed against all stored object embeddings to retrieve visually relevant objects.
\end{enumerate}

The resulting candidates from both mechanisms are combined into a unified set of relevant video frames.
For objects retrieved via embedding similarity, their corresponding semantic and relational context is extracted from the knowledge graph, yielding the same structured representation as graph-based retrieval.
This produces a unified subgraph that integrates symbolic relationships with visually grounded object cues.
We evaluate two SigLIP2 model sizes---\emph{GER-L} (Large) and \emph{GER-G} (Giant)---in \Cref{sec:experiments}.

\section{Benchmarks and Evaluation Protocol}
\label{sec:benchmarks}
We evaluate VL-KnG on three complementary benchmarks spanning indoor and outdoor environments, all following the episodic-memory protocol of \Cref{sec:problem}: the video is observed once, and questions are answered afterwards.

\subsection{WalkieKnowledge Benchmark}
\begin{figure}[htbp]
    \centering
    \includegraphics[width=0.9\textwidth]{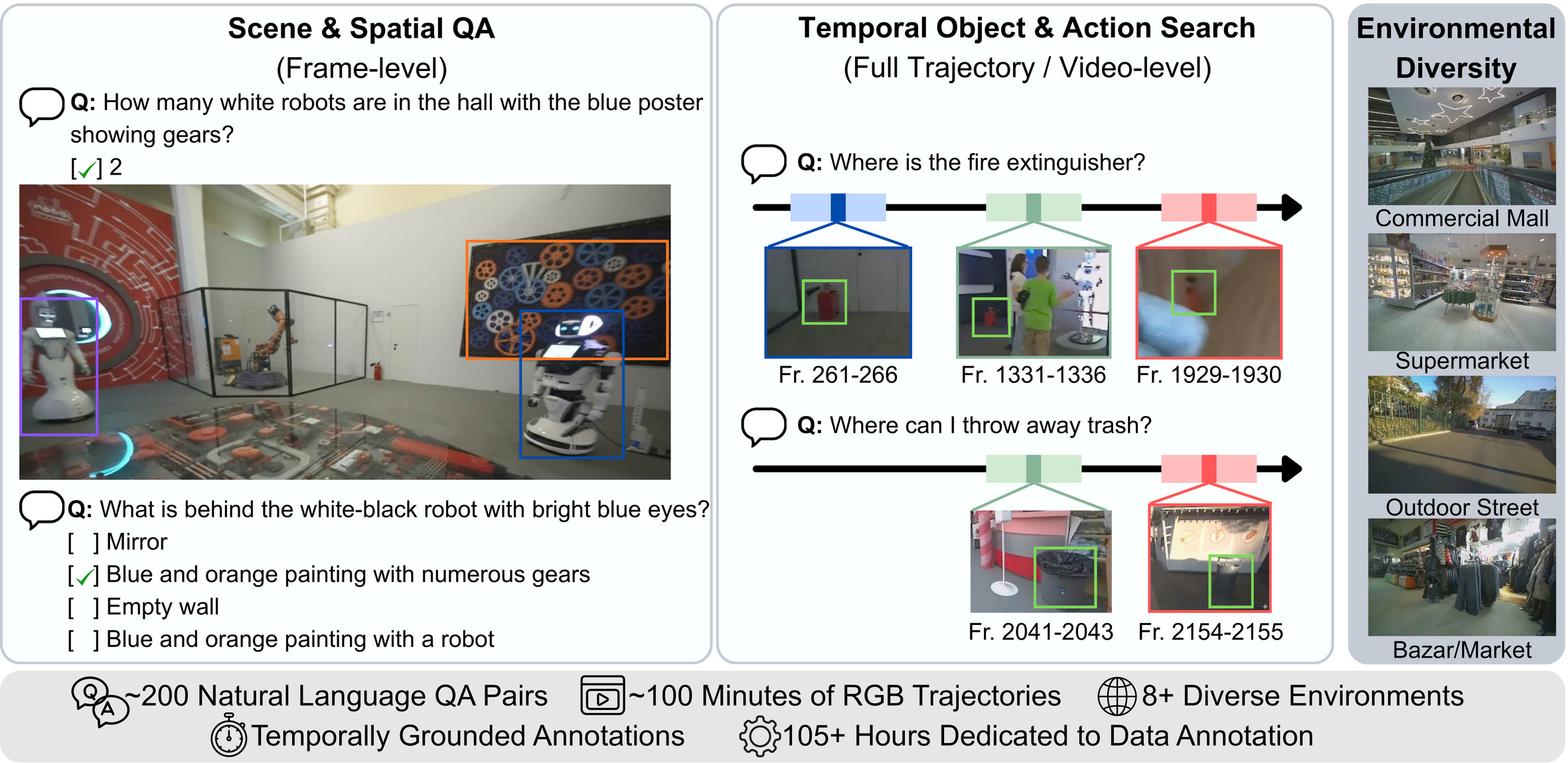}
    \caption{Overview of the WalkieKnowledge benchmark. The \textbf{left} panel shows scene \& spatial QA examples with frame-level bounding-box annotations and multiple-choice answers. The \textbf{center} panel presents temporal object \& action search queries requiring retrieval across full trajectories, with temporally grounded frame intervals. The \textbf{right} panel illustrates environmental diversity spanning commercial malls, supermarkets, outdoor streets, and bazaars. The benchmark comprises 193 manually annotated natural-language QA pairs over ${\sim}100$ minutes of egocentric RGB video across $8{+}$ diverse environments.}
    \label{fig:walkie-knowledge-overview}
\end{figure}

We introduce WalkieKnowledge, built on EgoWalk~\citep{akhtyamov2025egowalk}, spanning diverse indoor and outdoor environments (\cref{fig:walkie-knowledge-overview}).
It contains 8 egocentric trajectories, each recorded at a distinct geographic location, annotated with 193 questions across four types (object search, scene description, spatial relation, and action-place association), each linked to ground-truth frame intervals.

WalkieKnowledge fills a specific gap left by existing embodied QA benchmarks. OpenEQA~\citep{majumdar2024openeqa} evaluates short indoor scans where most questions are answerable from a handful of frames (32 frames per episode), and provides no retrieval-with-grounding metric; NaVQA~\citep{anwar2024remembr} evaluates robot navigation QA on sequences drawn from a single university campus, without joint frame-retrieval evaluation. WalkieKnowledge instead targets \emph{long, free-walking egocentric trajectories, each recorded in a geographically distinct environment} spanning diverse indoor and outdoor scenes (median 86 frames, ${\sim}$290 objects per scene), and evaluates retrieval and answer generation \emph{jointly} against frame-level ground truth.

\subsubsection{Frame Decimation}
We apply uniform frame decimation by sampling every $s$-th frame from each video ($s=60$ for WalkieKnowledge) for downstream visual processing and to facilitate meaningful ranking of relevant frames. To ensure fair evaluation, if no sampled frame falls within a question's annotated ground-truth frame ranges, we add at least one frame from those ranges containing the answer. This ensures all questions are covered without substantially increasing the total number of frames considered.

\subsubsection{Metrics}
Models are evaluated with retrieval and answer metrics. Retrieval Accuracy@$k$ checks whether a correct frame appears among the top-$k$ results, i.e., whether the system can locate the relevant moment in the video. Answer Accuracy is defined for multiple-choice questions, measuring whether the system selects the correct option. Additionally, we report Precision@$k$ (the proportion of relevant frames among the top $k$), Recall@$k$ (the proportion of relevant frames retrieved), and MRR@$k$ (the reciprocal rank of the first relevant frame among the top $k$).

\subsection{OpenEQA Benchmark}
OpenEQA~\citep{majumdar2024openeqa} contains 1,636 QA pairs across 180+ indoor environments, spanning seven categories.
We evaluate in the episodic-memory EQA (EM-EQA) setting, where the agent answers questions based solely on a stored sequence of past egocentric observations without further interaction with the environment. Performance is measured using \textbf{LLM-Match} (GPT-4 Turbo judges on a 1--5 scale, normalized to 0--100).

\subsection{NaVQA Benchmark}
NaVQA~\citep{anwar2024remembr} provides 210 QA pairs across 7 robot navigation sequences from CODa~\citep{zhang2024toward}, recorded in indoor and outdoor settings on a university campus.
We evaluate on descriptive (binary and text) queries, scoring binary questions by exact-match accuracy and text questions with an LLM judge, following the ReMEmbR protocol~\citep{anwar2024remembr}, and exclude query types that require metric 3D information (e.g., positional queries), which lies outside the scope of our monocular, reconstruction-free setting.

\section{Experiments}
\label{sec:experiments}
\subsection{Experimental Setup}
Our primary baselines are structured scene understanding methods that, like VL-KnG, build persistent representations from video. These include RoboHop~\citep{garg2024robohop} and WMNav~\citep{nie2025wmnav} on WalkieKnowledge; DAAAM~\citep{daaam2025arxiv}, ReMEmbR~\citep{anwar2024remembr}, and ConceptGraphs~\citep{gu2024conceptgraphs} on NaVQA; and GPT-4 w/ ConceptGraphs on OpenEQA. These methods share our goal of enabling persistent, queryable scene understanding and represent the most direct points of comparison. We additionally report frontier VLM baselines (Gemini~\citep{comanici2025gemini}, Qwen~\citep{bai2025qwen2}) to contextualize the accuracy--efficiency trade-off. Although these models achieve high accuracy through direct visual reasoning, they require re-processing video frames for every query, making them impractical for latency-sensitive or repeated-query deployment scenarios such as robotics.

We use Gemini~2.5~Flash~\citep{comanici2025gemini} for KG construction and reasoning.
We evaluate VL-KnG in the following configurations.
\textbf{Graph-based Retrieval (GR).} This configuration retrieves query-specific subgraphs from the knowledge graph containing the most semantically and spatially relevant objects and relationships, along with associated video frames. The retrieved subgraph is processed by the LLM to identify relevant frames and generate answers.
\textbf{Graph-Enhanced Retrieval (GER).}
Enhances GR by adding object-level visual similarity search over SigLIP2 embeddings~\citep{tschannen2025siglip}, built on top of graph-based retrieval (\Cref{sec:ger}).
We report results for two SigLIP2 model sizes:
\emph{GER-L} (Large) and \emph{GER-G} (Giant).
Detailed statistics of the constructed knowledge graphs across all benchmarks (objects per episode, unique spatial relations, and frames processed) are provided in Supplementary Table~S1.

\subsection{WalkieKnowledge Results}
WalkieKnowledge evaluates long-horizon scene understanding across four query categories, namely scene description, spatial relations, object search, and action--place association.
\Cref{tab:overall_performance} reports aggregate retrieval and answer-generation results, while \Cref{tab:perf_final_professional} provides a category-wise breakdown.
Graph-Enhanced Retrieval substantially improves over graph-based retrieval: GER-G reaches \textit{65.8\% Retrieval Acc.@1 and 0.735 MRR@5}, compared to 53.2\% and 0.572 for graph-only retrieval (GR). Even the graph-only configuration already outperforms the persistent-representation baseline RoboHop (34.7\% Retr.@1) and the open-weight Qwen~2.5~VL~72B (48.2\%) before any visual grounding is added. The adapted WMNav baseline attains retrieval accuracy comparable to GR (52.9\% vs.\ 53.2\% Retr.@1, and higher at @3/@5), but only by scoring every frame with a VLM at query time: lacking a persistent representation, it requires ${\sim}108$~s per query versus ${\sim}0.8$~s for VL-KnG---a ${\sim}135\times$ latency gap that directly illustrates the cost of re-processing observations for every question. With visual grounding, GER-G additionally surpasses WMNav on top-1 retrieval, precision, and ranking quality (65.8 vs.\ 52.9 Retr.@1; 0.735 vs.\ 0.654 MRR@5) at three orders of magnitude lower query latency. On ranking quality, VL-KnG approaches the strongest proprietary frontier VLMs (Gemini~2.5~Pro and Qwen3.5-Plus) while answering each query roughly \textit{30$\times$} faster (${\sim}0.8$~s vs.\ tens of seconds). We therefore position VL-KnG as offering competitive retrieval accuracy at substantially lower latency, rather than as uniformly surpassing frontier VLMs.
\Cref{tab:perf_final_professional} shows strong performance across categories. VL-KnG (GER-G) performs particularly well on scene description and action--place association queries, which benefit from persistent object identity and graph-based reasoning, while spatial relation queries remain more challenging without explicit 3D geometry.

\begin{table}[htbp]
\centering
\caption{Per-category performance breakdown on WalkieKnowledge (scene description, spatial relations, object search, and action--place association). All metrics are reported as percentages (\%), except Mean Reciprocal Rank (MRR). Higher is better ($\uparrow$). The top three results are highlighted by color as follows.
\protect\colorbox{best}{1st},
\protect\colorbox{second}{2nd}, and
\protect\colorbox{third}{3rd}.}
\label{tab:perf_final_professional}
\scriptsize
\setlength{\tabcolsep}{1.5pt}
\renewcommand{\arraystretch}{1.05}
\resizebox{\textwidth}{!}{%
\begin{tabular}{
  l @{\hskip 1em}
  S[table-format=1.2, round-mode=places, round-precision=2]
  S
  S[table-format=1.2, round-mode=places, round-precision=2]
  S
  S @{\hskip 1em}
  S[table-format=1.2, round-mode=places, round-precision=2]
  S
  S[table-format=1.2, round-mode=places, round-precision=2]
  S
  S @{\hskip 1em}
  S[table-format=1.2, round-mode=places, round-precision=2]
  S
  S[table-format=1.2, round-mode=places, round-precision=2]
  S @{\hskip 1em}
  S[table-format=1.2, round-mode=places, round-precision=2]
  S
  S[table-format=1.2, round-mode=places, round-precision=2]
  S
}
\toprule
\multirow{3}{*}{\textbf{Method}} & \multicolumn{5}{c}{\textbf{Scene Description}} & \multicolumn{5}{c}{\textbf{Spatial Relations}} & \multicolumn{4}{c}{\textbf{Object Search}} & \multicolumn{4}{c}{\textbf{Action-Place Assoc.}} \\
\cmidrule(lr){2-6} \cmidrule(lr){7-11} \cmidrule(lr){12-15} \cmidrule(lr){16-19}
& \rotatebox{90}{MRR@1$\uparrow$} & \rotatebox{90}{R@1$\uparrow$} & \rotatebox{90}{MRR@3$\uparrow$} & \rotatebox{90}{R@3$\uparrow$} & \rotatebox{90}{Acc$\uparrow$} &
  \rotatebox{90}{MRR@1$\uparrow$} & \rotatebox{90}{R@1$\uparrow$} & \rotatebox{90}{MRR@3$\uparrow$} & \rotatebox{90}{R@3$\uparrow$} & \rotatebox{90}{Acc$\uparrow$} &
  \rotatebox{90}{MRR@1$\uparrow$} & \rotatebox{90}{R@1$\uparrow$} & \rotatebox{90}{MRR@3$\uparrow$} & \rotatebox{90}{R@3$\uparrow$} &
  \rotatebox{90}{MRR@1$\uparrow$} & \rotatebox{90}{R@1$\uparrow$} & \rotatebox{90}{MRR@3$\uparrow$} & \rotatebox{90}{R@3$\uparrow$} \\
\midrule
RoboHop
& 0.27 & 16 & 0.34 & 29 & 24
& 0.31 & 19 & 0.40 & 37 & 29
& 0.37 & 21 & 0.44 & 45
& 0.42 & 19 & 0.53 & 36 \\
WMNav
& \third{0.61} & 31.11 & 0.71 & 49 & 58
& 0.18 & 17 & 0.44 & 30 & 45
& 0.61 & 41 & 0.77 & 73
& 0.62 & 31 & 0.58 & 45 \\
\midrule
Qwen2.5 VL 32B
& 0.35 & 11 & 0.51 & 42 & 41
& 0.31 & 18 & 0.37 & 28 & 43
& 0.37 & 22 & 0.57 & 54
& 0.26 & 12 & 0.48 & 36 \\
Qwen2.5 VL 72B
& 0.54 & 31 & 0.59 & 40 & 35
& 0.33 & 22 & 0.37 & 29 & 45
& 0.61 & 40 & 0.67 & 48
& 0.44 & 22 & 0.54 & 32 \\
\midrule
Qwen 3.5+
& \second{0.65} & \best{41} & 0.71 & \second{61} & \best{70}
& \second{0.55} & \second{37} & \second{0.65} & \third{59} & \best{63}
& \best{0.81} & \best{53} & \best{0.89} & \best{81}
& 0.62 & \third{32} & \third{0.70} & \third{53} \\
Gemini 2.5 Pro
& 0.59 & \third{33} & \second{0.73} & \third{60} & \second{68}
& \best{0.69} & \best{43} & \best{0.72} & 52 & \second{57}
& \second{0.77} & \second{51} & \second{0.87} & 65
& \second{0.66} & \second{33} & \best{0.76} & 50 \\
\midrule[\heavyrulewidth]
VL-KnG (GR)
& 0.57 & 28 & 0.60 & 50 & 54
& \second{0.55} & \third{33} & 0.57 & 49 & 47
& 0.57 & 32 & 0.60 & 55
& 0.44 & 19 & 0.50 & 42 \\
VL-KnG (GER-L)
& \second{0.65} & \second{34} & \third{0.72} & \second{61} & 54
& 0.51 & 31 & 0.60 & \second{60} & \third{49}
& 0.67 & 41 & 0.77 & \second{77}
& \third{0.64} & \second{33} & \second{0.73} & \second{61} \\
VL-KnG (GER-G)
& \best{0.68} & \best{41} & \best{0.74} & \best{66} & \third{60}
& \third{0.53} & \third{33} & \third{0.63} & \best{65} & 47
& \third{0.72} & \third{43} & \third{0.78} & \third{74}
& \best{0.70} & \best{38} & \best{0.76} & \best{64} \\

\bottomrule
\end{tabular}
}%
\end{table}

\begin{table}[htbp]
\centering
\scriptsize
\setlength{\tabcolsep}{2pt}
\sisetup{
  round-mode=places,
  round-precision=2,
  table-format=2.2,
  output-decimal-marker={.}
}
\caption{Overall performance of all evaluated methods on WalkieKnowledge.
All metrics are reported as percentages (\%), except Mean Reciprocal Rank (MRR). Higher is better ($\uparrow$); for latency, lower is better ($\downarrow$).
The top three results are highlighted by color as follows.
\protect\colorbox{best}{1st},
\protect\colorbox{second}{2nd}, and
\protect\colorbox{third}{3rd}. $^{\dagger}$Our re-implementation, adapted to this task. $^{\ddagger}$Adapted from its original object-goal navigation setting to monocular RGB frame retrieval. See ``On adapted baselines''.
}
\label{tab:overall_performance}
\resizebox{\textwidth}{!}{%
\begin{tabular}{l
S[table-format=2.2]
S[table-format=2.2]
S[table-format=2.2]
S[table-format=2.2]
S[table-format=2.2]
S[table-format=2.2]
S[table-format=2.2]
S[table-format=2.2]
S[table-format=2.2]}
\toprule
\multirow{2}{*}{\textbf{Metric}} &
\multicolumn{3}{c}{\textbf{VL-KnG}} &
{\multirow{2}{*}{\textbf{RoboHop}$^{\dagger}$}} &
{\multirow{2}{*}{\textbf{WMNav}$^{\ddagger}$}} &
\multicolumn{2}{c}{\textbf{Qwen 2.5 VL}} &
{\multirow{2}{*}{\textbf{Qwen 3.5+}}} &
{\multirow{2}{*}{\textbf{Gemini 2.5 Pro}}} \\
\cmidrule(lr){2-4} \cmidrule(lr){7-8}
& \textbf{GR}
& \textbf{GER-L}
& \textbf{GER-G}
& & &
\textbf{72B} & \textbf{32B}
& & \\
\midrule
\multicolumn{10}{l}{\textit{Retrieval Performance (\%)}} \\
\quad Retrieval Acc.@1$\uparrow$
& 53.16 & 61.66 & \third{65.80}
& 34.72 & 52.88 & 48.19 & 32.12 & \second{66.32} & \best{68.91} \\
\quad Retrieval Acc.@3$\uparrow$
& 62.11 & 80.31 & \third{81.35}
& 54.40 & 76.44 & 61.14 & 68.91 & \second{83.42} & \best{88.08} \\
\quad Retrieval Acc.@5$\uparrow$
& 64.21 & \third{85.49} & 83.94
& 62.69 & 84.44 & 61.14 & 70.47 & \second{87.56} & \best{89.12} \\
\quad Recall@1$\uparrow$
& 28.28 & 35.28 & \third{38.54}
& 19.28 & 30.94 & 28.93 & 16.38 & \best{41.15} & \second{40.94} \\
\quad Recall@3$\uparrow$
& 49.11 & \second{65.39} & \best{67.53}
& 37.50 & 62.62 & 37.64 & 40.56 & \third{64.34} & 56.97 \\
\quad Recall@5$\uparrow$
& 52.32 & \second{71.11} & \best{71.75}
& 47.40 & \third{70.47} & 37.64 & 42.09 & 69.78 & 58.13 \\
\quad Precision@1$\uparrow$
& 52.63 & 61.66 & \third{65.81}
& 35.75 & 52.87 & 48.19 & 32.13 & \second{66.32} & \best{68.91} \\
\quad Precision@3$\uparrow$
& 34.91 & \second{44.21} & \best{45.60}
& 24.35 & \third{41.18} & 21.94 & 25.22 & 40.59 & 34.54 \\
\quad Precision@5$\uparrow$
& 22.52 & \second{29.22} & \best{29.84}
& 18.55 & \third{28.50} & 13.16 & 15.85 & 27.67 & 21.56 \\
\midrule
\multicolumn{10}{l}{\textit{Ranking Quality}} \\
\quad MRR@1$\uparrow$
& 0.526 & 0.617 & \third{0.658}
& 0.347 & 0.528 & 0.482 & 0.321 & \second{0.663} & \best{0.689} \\
\quad MRR@3$\uparrow$
& 0.568 & 0.704 & \third{0.729}
& 0.434 & 0.631 & 0.542 & 0.485 & \second{0.744} & \best{0.778} \\
\quad MRR@5$\uparrow$
& 0.572 & 0.716 & \third{0.735}
& 0.454 & 0.654 & 0.542 & 0.489 & \second{0.754} & \best{0.781} \\
\midrule
\multicolumn{10}{l}{\textit{Generation Quality (\%)}} \\
\quad Answer Acc.
& 50.00 & 51.16 & \third{52.33}
& 26.74 & 50.59 & 40.70 & 41.86 & \best{66.28} & \second{61.63} \\
\midrule
\quad Latency (sec.)$\downarrow$
& \multicolumn{1}{c}{$\sim$ \textbf{0.8}} 
& \multicolumn{1}{c}{$\sim$ \textbf{0.8}} 
& \multicolumn{1}{c}{$\sim$ \textbf{0.8}}
& \multicolumn{1}{c}{\textemdash} 
& \multicolumn{1}{c}{$\sim 108$} 
& \multicolumn{1}{c}{\textemdash} 
& \multicolumn{1}{c}{\textemdash} 
& \multicolumn{1}{c}{$\sim 23$} 
& \multicolumn{1}{c}{$\sim 24$} \\
\bottomrule
\end{tabular}%
}
\end{table}

\paragraph{On adapted baselines.}
RoboHop has no publicly released implementation, so we re-implemented it from the paper. The ``performance optimizations'' noted in \Cref{tab:overall_performance} adapt it from its original visual-navigation setting to the natural-language-query~$\rightarrow$~frame-retrieval task evaluated here; this engineering was undertaken to make the comparison \emph{fair rather than weak}. WMNav is likewise designed for a different regime---active object-goal navigation with a VLM-driven world-model loop---and we adapted it to the offline retrieval task by bypassing its navigation, mapping, and action-selection components and reusing only its VLM interface: for each query, every decimated frame of the trajectory is assigned a graded relevance score via question-type-specific prompts, and frames are ranked by score with deterministic tie-breaking. This adaptation is accurate but expensive---because no persistent representation is built, every query triggers a full pass of per-frame VLM scoring, yielding the ${\sim}108$~s per-query latency reported in \Cref{tab:overall_performance}.

\subsection{OpenEQA Results}
\Cref{tab:openeqa_results} presents overall results on the OpenEQA benchmark with query latency.
GER-L achieves \textit{55.2}, outperforming GPT-4V (49.6) and GPT-4 w/ ConceptGraphs (36.5), indicating that structured knowledge graphs with explicit visual grounding provide more reliable reasoning than prior frame-sampling or graph-only approaches.
Incorporating visual grounding improves performance by \textit{+4.5 points} over graph-only retrieval (50.7 $\rightarrow$ 55.2), highlighting the importance of embedding-based object matching for accurate retrieval. We note that the larger SigLIP2-Giant encoder (GER-G, 54.7) does not strictly dominate the Large variant (GER-L, 55.2) here; consistent with the encoder-only ablation in \Cref{tab:siglip_only}, a larger visual encoder is not always better once retrieval is fused with graph reasoning, and the two operate within noise of each other on OpenEQA's short episodes.

\begin{table}[htbp]
\centering\small
\setlength{\tabcolsep}{4pt}
\caption{Effect of the visual encoder used \emph{without} the knowledge graph versus graph-only retrieval and the full GER-G system, on WalkieKnowledge (193 questions). The encoder-only rows rank KG-linked object crops by visual similarity and produce no answers (retrieval is not QA). All values in \% except MRR. Higher is better ($\uparrow$).}
\label{tab:siglip_only}
\begin{tabular}{@{}lccccc@{}}
\toprule
Method & Retr.@1$\uparrow$ & Retr.@3$\uparrow$ & Retr.@5$\uparrow$ & MRR@5$\uparrow$ & Ans.\ Acc.$\uparrow$\\
\midrule
SigLIP2-L (encoder only, no graph) & 62.18 & 75.65 & 81.87 & 0.695 & --\\
SigLIP2-G (encoder only, no graph) & 59.07 & 77.20 & 82.90 & 0.680 & --\\
GR (graph only, no SigLIP2)        & 53.16 & 62.11 & 64.21 & 0.572 & 50.00\\
\textbf{VL-KnG (GER-G)}            & \textbf{65.80} & \textbf{81.35} & \textbf{83.94} & \textbf{0.735} & \textbf{52.33}\\
\bottomrule
\end{tabular}
\end{table}

\begin{table}[htbp]
\centering
\caption{Results on OpenEQA (EM-EQA setting; 32 sampled frames per episode). LLM-Match scores an answer on a 1--5 scale against the ground truth, rescaled to 0--100 (higher is better); query latency is measured per question (lower is better). Human, GPT-4V, GPT-4 w/ ConceptGraphs, and Gemini~1.0 Pro Vision results are taken from the OpenEQA leaderboard.}
\label{tab:openeqa_results}
\scriptsize
\setlength{\tabcolsep}{0.5pt}
\begin{tabular}{lcc}
\toprule
\textbf{Method} & LLM-Match Score[\%] $\uparrow$  & Query Latency [sec] $\downarrow$\\
\midrule
Human & 86.8 & - \\
\midrule
GPT-4V & 49.6 & - \\
GPT-4 w/ ConceptGraphs & 36.5 & - \\
Gemini 1.0 Pro Vision & 44.9 & - \\
\midrule
Gemini 3 Flash & 76.8 & 10.5 \\
Qwen 3.5 Plus & 74.1 & $\sim$23 \\
Gemini 2.5 Flash & 69.8 & 6.8 \\
\midrule
VL-KnG (GR) & 50.7 & 0.8 \\
VL-KnG (GER-L) & 55.2 & 0.8 \\
VL-KnG (GER-G) & 54.7 & 0.8 \\
\bottomrule
\end{tabular}
\end{table}

\paragraph{Efficiency--accuracy trade-off.}
VL-KnG achieves roughly \textit{72\% of Gemini~3 Flash performance} while reducing query latency by an order of magnitude, with \textit{0.8~s per query vs.\ 6.8~s for Gemini~2.5 Flash and 10.5~s for Gemini~3 Flash}.
Notably, these speedups are measured on OpenEQA's short scenes (32 sampled frames per episode); for longer sequences such as WalkieKnowledge (up to 103 frames per scene), VLM latency grows proportionally while VL-KnG's remains roughly stable, widening the gap further.
This is because VL-KnG performs reasoning over a compact retrieved subgraph rather than processing all video frames, decoupling query latency from the raw video length.

Although VL-KnG does not match the best frontier VLMs on OpenEQA, this gap is expected given the benchmark's short episodes.
Most OpenEQA questions can be answered from a small set of frames, which favors end-to-end VLMs that directly attend to all images, whereas VL-KnG is designed for long-horizon video with many redundant frames, as in WalkieKnowledge (up to 103 frames per trajectory) and NaVQA (up to 431 frames per trajectory).
Representative success and failure cases on OpenEQA are shown in Supplementary Figure~S2.

\subsection{NaVQA Results}
\Cref{tab:navqa_results} presents NaVQA results.
GER-G achieves \textit{66.2\%} descriptive QA accuracy, the strongest among graph-based methods, approaching DAAAM (67.2\%) despite relying solely on monocular 2D knowledge graphs without depth sensing. Compared to graph-only (GR), GER-G improves descriptive QA accuracy by \textit{+34 percentage points}, demonstrating the importance of visual grounding for accurate answer generation.
Qualitative NavQA examples, including a case where the ground-truth annotation is itself ambiguous, are shown in Supplementary Figure~S3.

\begin{table}[htbp]
\centering
\caption{Results on the NaVQA descriptive question answering benchmark. Values are overall descriptive-question accuracy (0--1, higher is better); baseline results are taken from the respective papers.}
\label{tab:navqa_results}
\scriptsize
\setlength{\tabcolsep}{0.5pt}
\begin{tabular}{lcc}
\toprule
\textbf{Method} & \textbf{Descriptive Question Accuracy $\uparrow$} \\
\midrule
DAAAM (DAM-3B + GPT-5-mini) & \textbf{0.672} \\
ReMEmbR (NVILA-8B + GPT-5-mini) & 0.607 \\
ReMEmbR (NVILA-2B + GPT-5-mini) & 0.483 \\
ConceptGraphs & 0.299 \\
\midrule
VL-KnG (GR) & 0.324 \\
VL-KnG (GER-G) & \textbf{0.662} \\
\bottomrule
\end{tabular}
\end{table}

\subsection{Real-World Deployment}
To evaluate real-world deployment feasibility, we implemented VL-KnG on a differential-drive robot platform equipped with an Intel NUC11PHKI7C000 PC and an NVIDIA RTX 2060 GPU. The system uses SLAM Toolbox~\citep{macenski2021slam} and ROS Navigation Stack~\citep{guimaraes2016ros} for localization and navigation, with poses paired to source video frames for goal identification.

\paragraph{Deployment protocol.}
The robot first records a tour video of the environment, from which the spatiotemporal knowledge graph is built once offline; each graph node retains its source frame together with the corresponding SLAM pose. At query time, the top-1 retrieved frame defines the navigation goal. Its stored pose is sent to the ROS navigation stack, and a trial is counted as successful (binary) when the robot reaches that pose with the queried object in view. Evaluation covers queries spanning the four WalkieKnowledge question categories (object search, scene description, spatial relation, and action--place association) across real office environments.

The results are presented in \Cref{tab:hardware_results}.
At query time, the cost has two components: a lightweight lexical scoring pass over object descriptors, which grows with the number of graph objects but is computationally negligible, and LLM reasoning over the retrieved subgraph $\mathcal{G}_{\textit{sub}}$, which dominates latency and depends on the subgraph size rather than the video length. Since retrieval returns compact subgraphs ($|\mathcal{G}_{\textit{sub}}| \ll |\mathcal{G}|$) and the graph is reused across queries, average end-to-end query latency remains approximately $1$~s, versus roughly $24$~s for Gemini~2.5~Pro~\citep{comanici2025gemini}, while matching its success rate and answer accuracy (\Cref{tab:hardware_results}).

\begin{table}[htbp]
\centering
\caption{Real-world robot deployment results across the four WalkieKnowledge query categories in office environments. Success rate indicates that the robot reaches the goal pose with the queried object in view, while answer accuracy indicates the correctness of the generated answer.}
\label{tab:hardware_results}
\scriptsize
\setlength{\tabcolsep}{1.5pt}
\begin{tabular}{lcc}
\toprule
\textbf{Method} & \textbf{Success Rate (\%)} & \textbf{Answer Acc. (\%)} \\
\midrule
VL-KnG & 77.27 & 76.92 \\
Gemini 2.5 Pro & 77.27 & 76.92 \\
RoboHop & 27.27 & 23.08 \\
\bottomrule
\end{tabular}
\end{table}

\section{Ablation Studies}

\subsection{Retrieval Configurations and Impact of Visual Grounding}

In addition to the GR and GER configurations defined in \Cref{sec:experiments}, we evaluate two ablation configurations that isolate individual components, using Gemini~2.5 Flash for reasoning and frame localization throughout.

\textbf{Full Knowledge Graph (F).} This setting provides the entire knowledge graph as context to the LLM, enabling global reasoning across all available information but forgoing query-specific retrieval.

\textbf{Chunk-Wise Graph-based Retrieval (CwGR).} This setting queries iteratively across all local chunk graphs without cross-chunk associations, thereby isolating the contribution of spatiotemporal object association (STOA).

\Cref{tab:overall_performance_transposed} compares the four configurations. Removing cross-chunk association and treating chunks independently (CwGR) drops answer accuracy from $50.0$ to $37.2$ ($-12.8$ points), confirming that persistent identity---not retrieval alone---is what enables coherent long-horizon reasoning. Providing the full graph as context (F) reaches $57.5\%$ Retr.@1 and the highest answer accuracy overall ($58.1\%$), confirming that global graph context aids answer generation; however, its per-query context grows with the full graph size, whereas retrieval passes a compact, bounded subgraph, and GER-G surpasses F on every retrieval and ranking metric ($65.8$ vs.\ $57.5$ Retr.@1; $73.5$ vs.\ $62.8$ MRR@5). Subgraph retrieval thus trades a modest amount of answer accuracy for scalable per-query cost, and recovers superior frame localization once visual grounding is added---the property that matters for goal identification, where the top-ranked frame defines the navigation target.
Furthermore, Graph-Enhanced Retrieval (GER) consistently improves over GR by incorporating visually grounded object retrieval, indicating that structured graph reasoning and object-level visual similarity are complementary for frame localization and ranking.

\begin{table}[htbp]
\centering
\sisetup{
  round-mode=places,
  round-precision=1,
  table-format=2.1,
  output-decimal-marker={.}
}
\caption{Comparison of VL-KnG retrieval configurations on WalkieKnowledge. The progression from CwGR (no cross-chunk association) to GR (with association) to GER (with visual grounding) isolates the contribution of each component. All metrics are reported as percentages (\%). Higher is better ($\uparrow$).}

\resizebox{\textwidth}{!}{
\begin{tabular}{l
S S S
S S S
S S S
S S S
S}
\toprule
\multirow{2}{*}{\textbf{Method}}
& \multicolumn{3}{c}{\textbf{Retr. Acc. $\uparrow$}}
& \multicolumn{3}{c}{\textbf{Recall $\uparrow$}}
& \multicolumn{3}{c}{\textbf{Precision $\uparrow$}}
& \multicolumn{3}{c}{\textbf{MRR $\uparrow$}}
& {\multirow{2}{*}{\textbf{Ans. Acc. $\uparrow$}}} \\
\cmidrule(lr){2-4} \cmidrule(lr){5-7} \cmidrule(lr){8-10} \cmidrule(lr){11-13}
& {@1} & {@3} & {@5}
& {@1} & {@3} & {@5}
& {@1} & {@3} & {@5}
& {@1} & {@3} & {@5}
&  \\
\midrule
VL-KnG (F) & 57.51 & 69.43 & 69.95 & 27.9 & 53.7 & 57.3 & 57.51 & 39.55 & 25.80 & 57.51 & 62.69 & 62.82 & 58.14 \\
VL-KnG (CwGR) & 50.78 & 56.99 & 57.51 & 28.67 & 44.51 & 45.42 & 48.70 & 30.05 & 18.76 & 48.70 & 52.33 & 52.43 & 37.21 \\
VL-KnG (GR) & 53.16 & 62.11 & 64.21 & 28.28 & 49.11 & 52.32 & 52.63 & 34.91 & 22.52 & 52.63 & 56.75 & 57.23 & 50.00 \\
\midrule[\heavyrulewidth]
VL-KnG (GER-L)
& 61.66 & 80.31 & 85.49
& 35.3 & 65.4 & 71.1
& 61.7 & 44.2 & 29.2
& 61.7 & 70.4 & 71.6
& 51.16 \\
VL-KnG (GER-G)
& 65.80 & 81.35 & 83.94
& 38.5 & 67.5 & 71.8
& 65.8 & 45.6 & 29.8
& 65.8 & 72.9 & 73.5
& 52.33 \\
\bottomrule
\end{tabular}
}
\label{tab:overall_performance_transposed}
\end{table}

A natural concern is whether these gains come from the SigLIP2 visual encoder alone rather than from the knowledge graph. \Cref{tab:siglip_only} answers this directly by comparing the visual encoder used \emph{without} the graph (ranking KG-linked object crops purely by visual similarity) against graph-only retrieval and the full GER-G system. Three observations follow. First, full GER-G improves over its matched encoder-only counterpart (SigLIP2-G) by $+6.7$ Retrieval Acc.@1 and $+5.5$ MRR@5 points, and exceeds the best encoder-only configuration on every metric, so the graph adds measurable retrieval gain on top of the visual encoder. Second, the encoder alone produces no answers at all (zero answer accuracy), meaning it can only rank object instances, whereas all of VL-KnG's $50$--$52\%$ answer accuracy comes from graph-grounded LLM reasoning. Third, graph-only GR ($53.2\%$ Retr.@1) already exceeds RoboHop and the open-weight Qwen~2.5~VL~72B before any visual grounding is added, and matches the adapted WMNav baseline at a small fraction of its per-query cost. The two components are therefore complementary rather than redundant. The knowledge graph supplies persistent object memory, temporal linkage, and relational reasoning, while SigLIP2 sharpens ranking over those persistent instances.

\subsection{Effect of LLM Backbone and Decoding Temperature}

\Cref{tab:ablation_llm_temp} compares VL-KnG under different LLM backbones and decoding temperatures, with and without SigLIP2-Giant visual grounding.
Within the graph-only GR pipeline, Qwen3.5-Plus at lower temperature ($T{=}0.1$) yields slightly lower retrieval metrics but higher answer accuracy than $T{=}0.6$, reflecting a trade-off between exploration in retrieval and stability in answer selection.
Across backbones, Gemini~2.5 Flash achieves stronger retrieval and ranking (Retr.@1/@3/@5 and MRR) than Qwen3.5-Plus, while Qwen3.5-Plus can match or exceed answer accuracy in some configurations.
Adding visual grounding (GER-G) improves retrieval metrics substantially for both backbones.
For Gemini~2.5 Flash, knowledge graph construction is performed at $T{=}0.1$ to encourage stable structured outputs, while GraphRAG-based retrieval and answer generation use $T{=}0.7$.
A per-question-type breakdown of this ablation is provided in Supplementary Table~S2.

\begin{table}[htbp]
  \centering
  \sisetup{round-mode=places, round-precision=1, table-format=2.1, output-decimal-marker={.}}
  \caption{Effect of the LLM backbone (Gemini~2.5 Flash, G2.5F, vs.\ Qwen3.5-Plus, Q3.5P) and decoding temperature $T$ on WalkieKnowledge, within the graph-only (GR) and visually grounded (GER-G) pipelines. All values in \%. Higher is better ($\uparrow$).}
  \scriptsize
  \label{tab:ablation_llm_temp}
  \resizebox{\textwidth}{!}{%
  \begin{tabular}{l
  S S S
  S S S
  S S S
  S S S
  S}
  \toprule
  \multirow{2}{*}{\textbf{Setting}}
  & \multicolumn{3}{c}{\textbf{Retr. Acc. $\uparrow$}}
  & \multicolumn{3}{c}{\textbf{Recall $\uparrow$}}
  & \multicolumn{3}{c}{\textbf{Precision $\uparrow$}}
  & \multicolumn{3}{c}{\textbf{MRR $\uparrow$}}
  & {\multirow{2}{*}{\textbf{Ans. Acc. $\uparrow$}}} \\
  \cmidrule(lr){2-4} \cmidrule(lr){5-7} \cmidrule(lr){8-10} \cmidrule(lr){11-13}
  & {@1} & {@3} & {@5}
  & {@1} & {@3} & {@5}
  & {@1} & {@3} & {@5}
  & {@1} & {@3} & {@5}
  & \\
  \midrule
  VL-KnG (GR), Q3.5P, $T{=}0.1$
  & 44.04 & 52.33 & 52.85
  & 26.81 & 42.96 & 45.98
  & 44.04 & 29.02 & 19.27
  & 44.04 & 47.84 & 47.97
  & 59.30 \\
  VL-KnG (GR), Q3.5P, $T{=}0.6$
  & 44.56 & 56.99 & 58.03
  & 26.44 & 45.87 & 48.47
  & 44.56 & 30.74 & 19.90
  & 44.56 & 50.43 & 50.69
  & 53.49 \\
  VL-KnG (GR), G2.5F
  & 53.16 & 62.11 & 64.21
  & 28.28 & 49.11 & 52.32
  & 52.63 & 34.91 & 22.52
  & 52.63 & 56.75 & 57.23
  & 50.00 \\
  \midrule
  VL-KnG (GER-G), Q3.5P, $T{=}0.6$
  & 59.59 & 77.20 & 82.90
  & 35.20 & 64.30 & 70.82
  & 59.58 & 42.14 & 29.53
  & 59.58 & 67.19 & 68.53
  & 56.98 \\
  VL-KnG (GER-G), G2.5F
  & 65.80 & 81.35 & 83.94
  & 38.50 & 67.50 & 71.80
  & 65.80 & 45.60 & 29.80
  & 65.80 & 72.90 & 73.50
  & 52.33 \\
  \bottomrule
  \end{tabular}%
  }
\end{table}

\subsection{Chunk Size for KG Construction}
\label{sec:chunk_size_ablation}

\Cref{tab:chunk_size_ablation} evaluates the effect of the chunk size $b$ (number of frames per VLM call) using graph-only retrieval (GR). The default $b{=}8$ used throughout all experiments is compared against $b{=}5$ and $b{=}15$ on the two longest WalkieKnowledge trajectories (84 questions in total), focusing on a high-question-count setting.

\begin{table}[htbp]
  \centering
  \sisetup{
    round-mode=places,
    round-precision=1,
    table-format=2.1,
    output-decimal-marker={.}
  }
  \caption{Chunk size ablation on the two longest WalkieKnowledge trajectories (GraphRAG mode, 84 questions). All values in \%. Higher is better ($\uparrow$).}
  \label{tab:chunk_size_ablation}
  \resizebox{\textwidth}{!}{
  \begin{tabular}{c
  S S S
  S S S
  S S S
  S S S
  S}
  \toprule
  \multirow{2}{*}{\textbf{Chunk size} $b$}
  & \multicolumn{3}{c}{\textbf{Retr.\ Acc.\ $\uparrow$}}
  & \multicolumn{3}{c}{\textbf{Recall $\uparrow$}}
  & \multicolumn{3}{c}{\textbf{Precision $\uparrow$}}
  & \multicolumn{3}{c}{\textbf{MRR $\uparrow$}}
  & {\multirow{2}{*}{\textbf{Ans.\ Acc.\ $\uparrow$}}} \\
  \cmidrule(lr){2-4} \cmidrule(lr){5-7} \cmidrule(lr){8-10} \cmidrule(lr){11-13}
  & {@1} & {@3} & {@5}
  & {@1} & {@3} & {@5}
  & {@1} & {@3} & {@5}
  & {@1} & {@3} & {@5}
  & \\
  \midrule
  5 & 30.95 & 36.90 & 39.29 & 13.47 & 25.68 & 30.60 & 29.76 & 22.22 & 17.38 & 29.76 & 33.14 & 33.73 & 46.15 \\
  \textbf{8 (default)} & \textbf{46.43} & \textbf{51.19} & \textbf{52.38} & \textbf{21.04} & \textbf{35.56} & \textbf{38.53} & \textbf{46.42} & \textbf{30.16} & \textbf{20.36} & \textbf{46.43} & \textbf{48.81} & \textbf{49.11} & \textbf{61.54} \\
  15 & 27.38 & 42.86 & 44.05 & 10.03 & 28.55 & 31.66 & 26.19 & 23.81 & 17.86 & 26.19 & 34.33 & 34.62 & 53.85 \\
  \bottomrule
  \end{tabular}
  }
\end{table}

The chunk size has a strong impact on the quality of the constructed knowledge graph, and the default $b{=}8$ consistently achieves the best performance across all retrieval metrics.
The degradation at smaller chunks ($b{=}5$) suggests that insufficient temporal context harms graph construction because fewer frames per chunk split object observations across more segments. This reduces the likelihood that repeated sightings are consolidated into stable nodes and relations, and the resulting fragmentation propagates to retrieval.
Conversely, larger chunks ($b{=}15$) also degrade performance because the resulting graph segments become denser and introduce more competing entities and relations within a single chunk, increasing ambiguity during retrieval.
The intermediate setting $b{=}8$ captures enough temporal continuity to consolidate observations while keeping each graph segment sufficiently focused, producing the most consistent retrieval behavior and the highest end-to-end QA accuracy.

\section{Computational Trade-offs}
\label{sec:when_to_use}

\begin{figure}[htbp]
    \centering
    \includegraphics[width=\textwidth]{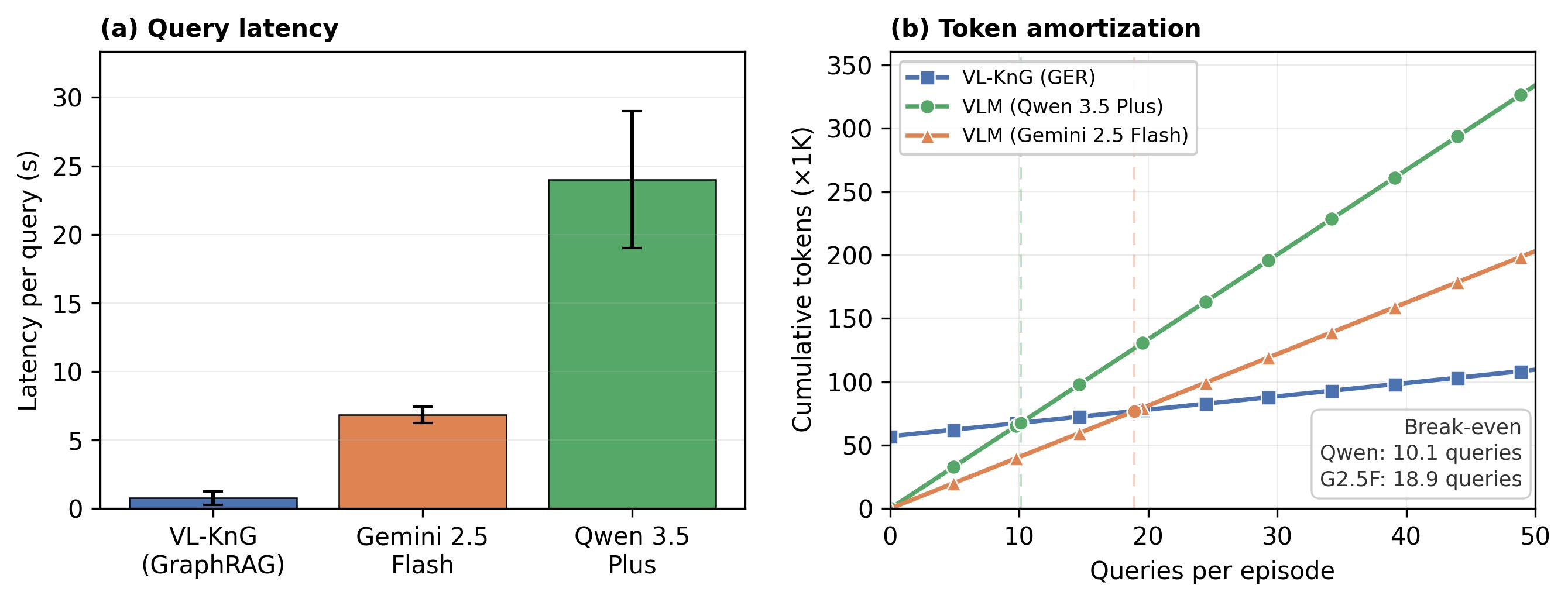}
    \caption{Efficiency comparison on OpenEQA. 
(a) Mean per-query latency with standard deviation ($n{=}50$ for VL-KnG, $n{=}10$ per VLM baseline), comparing VL-KnG (GraphRAG) with Gemini~2.5~Flash and Qwen~3.5~Plus.
(b) Cumulative token usage per episode against both VLM baselines, showing that VL-KnG pays an upfront graph-construction cost and then amortizes it as more questions are asked about the same scene.}
    \label{fig:cost}
\end{figure}

\Cref{fig:cost} analyzes the computational trade-off between VL-KnG and direct VLM querying on OpenEQA (episodes 000--004, 50 questions).
Per-query latency ($n{=}50$ questions for VL-KnG; $n{=}10$ per VLM baseline due to inference cost) is roughly $9\times$ lower than Gemini~2.5~Flash (${\sim}0.8$\,s vs.\ ${\sim}6.8$\,s) and ${\sim}30\times$ lower than Qwen3.5-Plus (${\sim}23$\,s), because queries are answered from a pre-built text-only knowledge graph rather than requiring image processing at inference time.
The token amortization curve shows cumulative token usage as additional questions are asked.
VL-KnG incurs an upfront cost to construct the knowledge graph but breaks even after ${\sim}10$ queries per episode against Qwen3.5-Plus and ${\sim}19$ queries against Gemini~2.5~Flash, after which the token gap grows approximately linearly.

\paragraph{Latency protocol.} The latency numbers reported in \Cref{tab:overall_performance} are computed as follows. For VL-KnG, the knowledge graph is constructed once per trajectory, after which questions are processed sequentially; latency is measured per query using the pre-built graph, reflecting repeated querying of the same scene.
In contrast, direct VLM baselines receive the entire video trajectory together with all questions in a single run, and the reported latency is the total runtime divided by the number of questions. This amortizes video ingestion across all questions and therefore \emph{favors} the VLM baselines; if an additional question is introduced later, the VLM must re-process the full trajectory, whereas VL-KnG reuses the existing graph and performs only retrieval and answer generation. WMNav latency is measured per query, as its adaptation scores each frame independently for every question.

\section{Conclusion}
\label{sec:conclusion}
We presented VL-KnG, a training-free framework that constructs persistent spatiotemporal knowledge graphs from monocular egocentric video, enabling structured and explainable scene understanding.
By decoupling knowledge construction from query processing, VL-KnG provides persistent memory and interpretable reasoning while avoiding the poor scaling of direct VLM inference with video duration.
Evaluation across three benchmarks shows that VL-KnG achieves accuracy competitive with frontier VLMs on embodied scene understanding while answering queries at substantially lower latency, and that it surpasses prior persistent-representation baselines and open-weight VLMs in several settings.
Graph-Enhanced Retrieval, combining structured subgraph reasoning with visual grounding, plays a key role in bridging the gap between text-level graph representations and frame-level localization.
Real-world robot deployment further demonstrates practical applicability, with query latency remaining stable as the observation history grows.

\noindent\textbf{Limitations.} Fine-grained spatial reasoning remains a limitation, as the knowledge graph encodes positions at the frame level without explicit 3D geometry. 
KG construction quality is also bounded by the underlying VLM's detection capability. 
Finally, the current STOA module assumes relatively static scenes where objects do not frequently change state or location, which may limit performance in highly dynamic environments.
Future work will explore dynamic environments with changing object states, integration of multi-modal sensing for richer graph construction, and scaling VL-KnG to long-duration video streams spanning hours of continuous observation. 
More broadly, our results suggest that persistent structured representations offer a promising path toward scalable and explainable embodied scene understanding beyond direct end-to-end VLM inference.

\section*{Conflict of Interest Statement}
The authors declare that the research was conducted in the absence of any commercial or financial relationships that could be construed as a potential conflict of interest.

\section*{Author Contributions}
Conceptualization, M.A.M., S.Z., S.L. and G.F.; methodology, M.A.M.; software, M.A.M. and S.L.; validation, M.A.M. and S.L.; formal analysis, M.A.M. and S.L.; investigation, M.A.M., S.L., T.A. and D.N.; resources, G.F.; data curation, S.L., A.N. and M.A.M.; writing---original draft preparation, M.A.M.; writing---review and editing, S.L., T.A., A.N., S.Z. and G.F.; visualization, M.A.M., S.L., T.A., D.N. and A.N.; project administration, M.A.M.; supervision, S.Z. and G.F. All authors have read and agreed to the submitted version of the manuscript.


\section*{Data Availability Statement}
The source code and data will be made publicly available upon acceptance.

\section*{Supplementary Material}
The Supplementary Material is included at the end of this preprint. It contains a step-by-step visualization of knowledge graph construction (Supplementary Figure~S1), detailed knowledge graph statistics (Supplementary Table~S1), a per-question-type breakdown of the backbone ablation (Supplementary Table~S2), qualitative examples on OpenEQA and NavQA (Supplementary Figures~S2--S3), and the complete prompt templates used across all experiments.

\bibliographystyle{Frontiers-Harvard}
\bibliography{references}\clearpage
\begingroup
\setcounter{section}{0}
\setcounter{figure}{0}
\setcounter{table}{0}
\renewcommand{\thesection}{S\arabic{section}}
\renewcommand{\thefigure}{S\arabic{figure}}
\renewcommand{\thetable}{S\arabic{table}}
\renewcommand{\figurename}{Supplementary Figure}
\renewcommand{\tablename}{Supplementary Table}
\crefname{figure}{Supplementary Figure}{Supplementary Figures}
\crefname{table}{Supplementary Table}{Supplementary Tables}
\Crefname{figure}{Supplementary Figure}{Supplementary Figures}
\Crefname{table}{Supplementary Table}{Supplementary Tables}
\makeatletter
\def\fnum@table{Supplementary Table~\thetable}
\def\fnum@figure{Supplementary Figure~\thefigure}
\makeatother
\crefname{section}{Supplementary Section}{Supplementary Sections}
\Crefname{section}{Supplementary Section}{Supplementary Sections}

\begin{center}
{\helveticabold\fontsize{20}{22}\selectfont Supplementary Material\par}
\end{center}
\vspace{1em}
\noindent This Supplementary Material accompanies the main article above.
It provides a step-by-step visualization of knowledge graph construction (Section~\ref{ssec:stoa_supp}), detailed knowledge graph statistics (Section~\ref{ssec:kg_stats_supp}), a per-question-type breakdown of the LLM backbone ablation (Section~\ref{ssec:ablation_supp}), qualitative examples (Section~\ref{ssec:qual_supp}), and the complete prompt templates used across all experiments (Section~\ref{ssec:prompts_supp}).

\section{Supplementary Methods}

\subsection{Pipeline Steps Visualization}
\label{ssec:stoa_supp}

\Cref{fig:pipeline_steps} illustrates the three stages of VL-KnG's knowledge graph construction on a single episode: (a)~raw input frames sampled from the egocentric video, (b)~VLM-based object detection with bounding boxes and tracked IDs, and (c)~the resulting structured knowledge graph with object attributes and spatial relationships.

\begin{figure}[h]
    \centering
    \includegraphics[width=\textwidth]{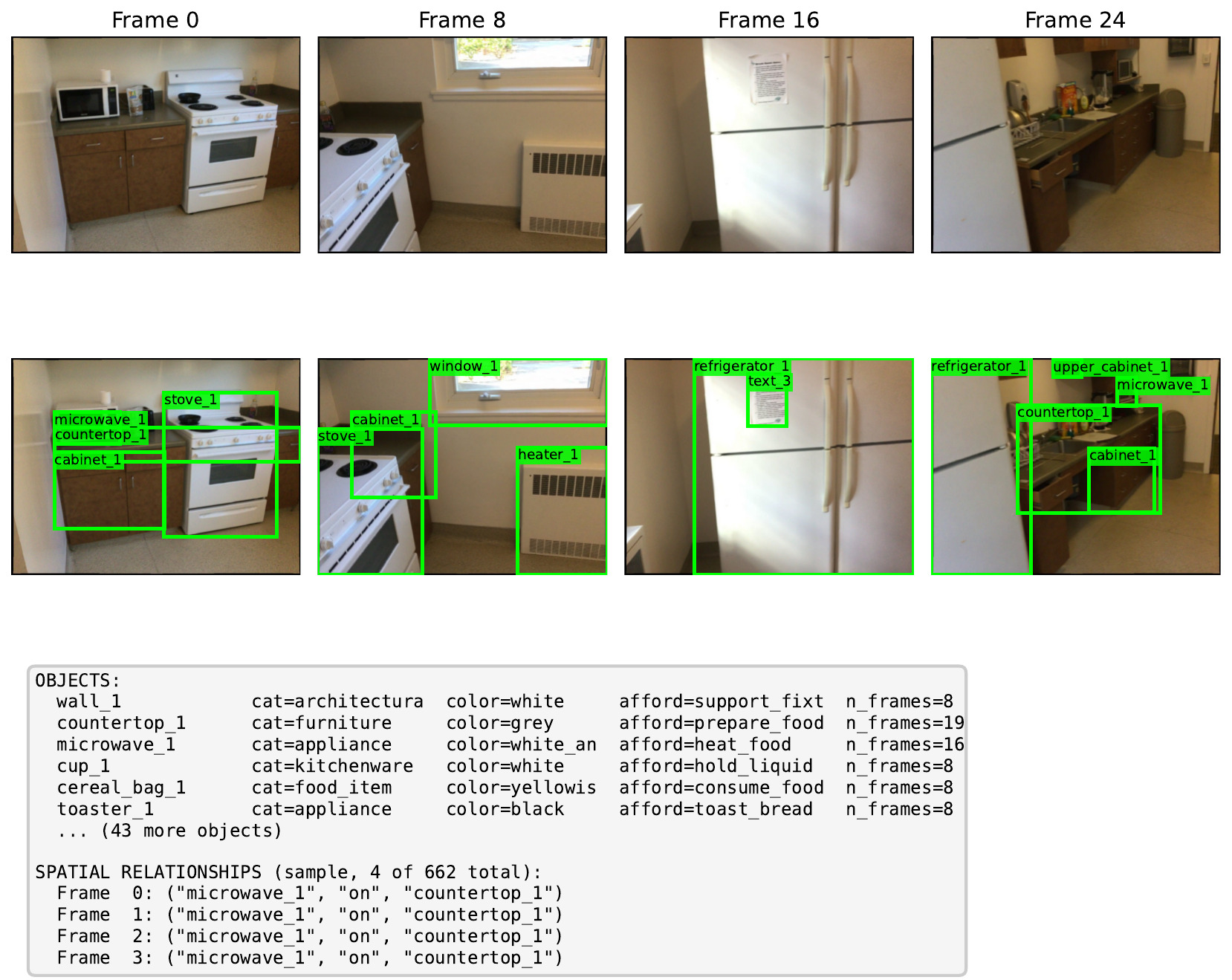}
    \caption{VL-KnG pipeline stages for one episode. (a)~Sampled input frames. (b)~Object detection with bounding boxes and cross-frame ID tracking. (c)~Structured knowledge graph output with object attributes and spatial relationships.}
    \label{fig:pipeline_steps}
\end{figure}

\section{Knowledge Graph Statistics}
\label{ssec:kg_stats_supp}

\Cref{tab:kg_stats_table} provides detailed knowledge graph statistics across OpenEQA, WalkieKnowledge, and NaVQA, computed from all built KGs.

\begin{table}[h]
\centering
\small
\caption{Knowledge graph statistics across benchmarks. Values show median (Q1--Q3). Unique spatial relations count each \texttt{(subject, relation, object)} triple once regardless of how many frames it appears in.}
\label{tab:kg_stats_table}
\begin{tabular}{l cccc}
\toprule
\textbf{Metric} & \textbf{OpenEQA} & \textbf{OpenEQA} & \textbf{WalkieKnow.} & \textbf{NaVQA} \\
& \textbf{ScanNet} (89) & \textbf{HM3D} (63) & (8) & (7) \\
\midrule
Objects per episode & 38 (28--48) & 51 (39--67) & 290 (250--450) & 287 (244--316) \\
Unique spatial relations & 83 (67--101) & 105 (85--131) & 704 (477--847) & 1{,}072 (618--1{,}275) \\
Frames processed & 32 (32--32) & 32 (32--32) & 86 (44--90) & 265 (152--334) \\
\bottomrule
\end{tabular}
\end{table}

HM3D episodes produce larger knowledge graphs (median 51 objects) than ScanNet episodes (median 38), consistent with HM3D's larger multi-room environments.
WalkieKnowledge and NaVQA sequences produce substantially larger graphs (median 290 and 287 objects respectively), reflecting the richer visual complexity of shopping malls and the continuous observation of diverse objects along robot navigation trajectories.
The object distributions clearly reflect each domain: OpenEQA is dominated by indoor furniture and fixtures, WalkieKnowledge by signage, clothing, and retail fixtures, while NaVQA features pedestrians, vehicles, and urban infrastructure.

\section{Per-Question-Type Ablation}
\label{ssec:ablation_supp}

\Cref{tab:ablation_per_type} breaks down the LLM backbone and visual grounding ablation of the main article by WalkieKnowledge question type, reporting retrieval accuracy (Retr.@$k$), mean reciprocal rank (MRR@1), recall (R@3), and multiple-choice answer accuracy where applicable.

\begin{table}[h]
\centering
\caption{Per-question-type breakdown of the LLM backbone and visual grounding ablation on WalkieKnowledge. Retr.@$k$: retrieval accuracy; MRR@1: mean reciprocal rank; R@3: recall at 3. All values in \%.}
\label{tab:ablation_per_type}
\resizebox{\textwidth}{!}{%
\begin{tabular}{l ccccc ccccc ccccc}
\toprule
\multirow{2}{*}{\textbf{Question type}}
& \multicolumn{5}{c}{Qwen3.5 Plus $T{=}0.1$}
& \multicolumn{5}{c}{Qwen3.5 Plus $T{=}0.6$}
& \multicolumn{5}{c}{+ SigLIP2 Giant} \\
\cmidrule(lr){2-6} \cmidrule(lr){7-11} \cmidrule(lr){12-16}
& Retr.@1$\uparrow$ & Retr.@3$\uparrow$ & Retr.@5$\uparrow$ & MRR@1$\uparrow$ & R@3$\uparrow$
& Retr.@1$\uparrow$ & Retr.@3$\uparrow$ & Retr.@5$\uparrow$ & MRR@1$\uparrow$ & R@3$\uparrow$
& Retr.@1$\uparrow$ & Retr.@3$\uparrow$ & Retr.@5$\uparrow$ & MRR@1$\uparrow$ & R@3$\uparrow$ \\
\midrule
Object search
& 44.6 & 51.8 & 53.6 & 44.6 & 44.5
& 49.1 & 61.4 & 63.2 & 49.1 & 51.2
& 64.9 & 82.5 & 84.2 & 64.9 & 69.7 \\
Scene description
& 43.2 & 56.8 & 56.8 & 43.2 & 47.1
& 43.2 & 48.7 & 51.4 & 43.2 & 40.7
& 59.5 & 78.4 & 94.6 & 59.5 & 65.7 \\
Action-place association
& 52.1 & 60.4 & 60.4 & 52.1 & 42.8
& 44.0 & 60.0 & 60.0 & 44.0 & 41.2
& 66.0 & 80.0 & 82.0 & 66.0 & 60.5 \\
Spatial relation
& 38.8 & 44.9 & 44.9 & 38.8 & 40.8
& 40.8 & 55.1 & 55.1 & 40.8 & 48.3
& 46.9 & 67.3 & 73.5 & 46.9 & 60.8 \\
\midrule
\multicolumn{16}{l}{\textit{Answer accuracy (multiple choice)}} \\
Scene description
& \multicolumn{5}{c}{59.46} & \multicolumn{5}{c}{43.24} & \multicolumn{5}{c}{45.95} \\
Spatial relation
& \multicolumn{5}{c}{59.18} & \multicolumn{5}{c}{61.22} & \multicolumn{5}{c}{65.31} \\
\bottomrule
\end{tabular}%
}
\end{table}

\section{Qualitative Examples}
\label{ssec:qual_supp}

\subsection{OpenEQA}

\Cref{fig:qualitative} shows representative success and failure cases comparing VL-KnG (GER-Giant) against a direct VLM baseline (Gemini~3 Flash) on OpenEQA. Scores are LLM-Match (1--5 scale, GPT-4o-mini judge). VL-KnG succeeds on questions where its structured attribute storage (object states, materials, affordances) directly provides the answer. It fails when the question requires precise spatial reasoning or relies on visual details not captured in the graph's text encoding.

\begin{figure}[h]
    \centering
    \includegraphics[width=\textwidth]{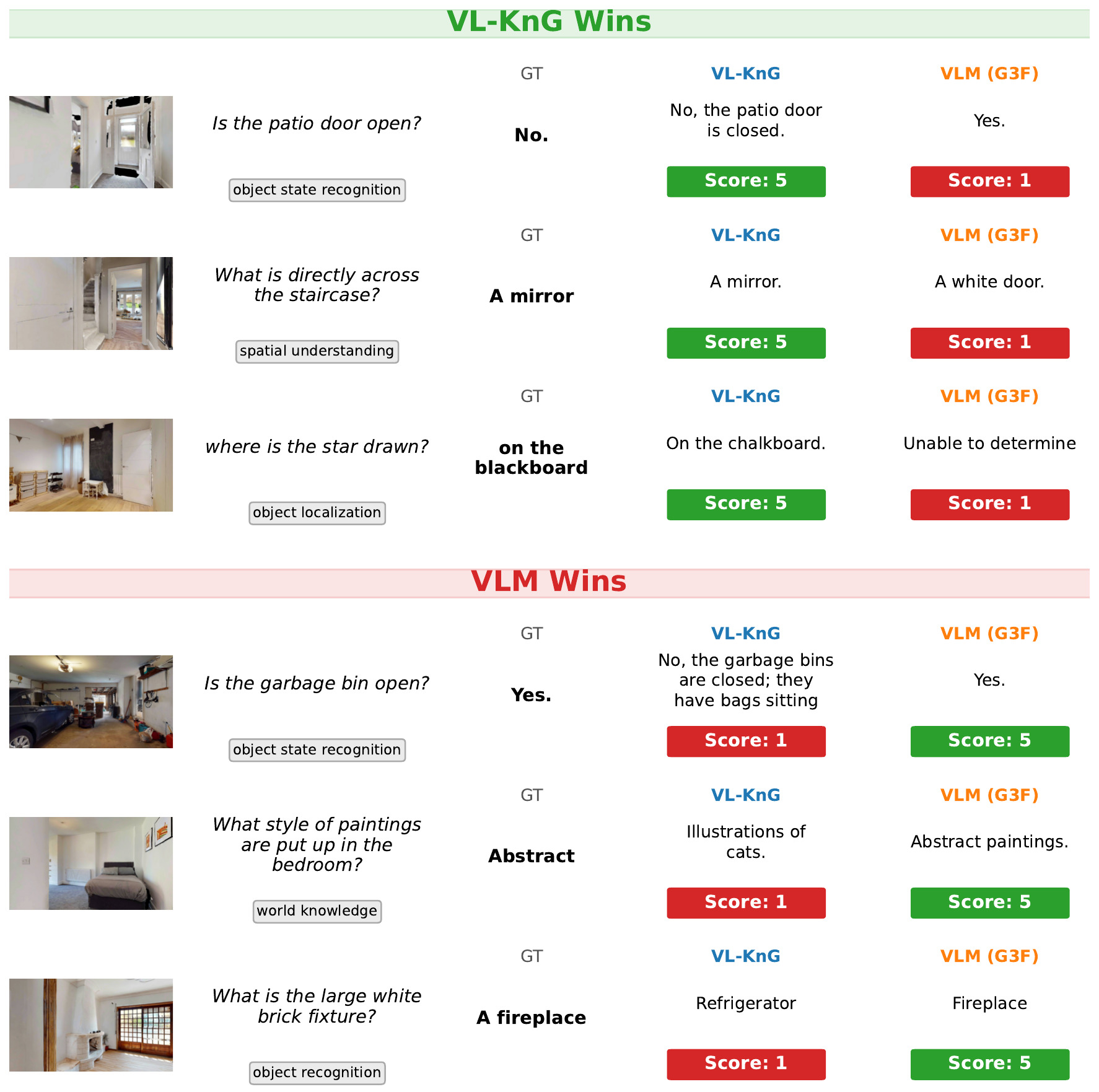}
    \caption{Qualitative comparison of VL-KnG vs.\ VLM (Gemini~3 Flash) on OpenEQA. \textbf{Top:} VL-KnG wins---graph-stored attributes answer the question; the VLM guesses incorrectly. \textbf{Bottom:} VLM wins---direct visual inspection outperforms the KG's text-level encoding. The frame shown is the most relevant episode frame identified by object keyword matching.}
    \label{fig:qualitative}
\end{figure}

\subsection{NaVQA}

\Cref{fig:navqa_qual} shows qualitative examples from the NaVQA benchmark, where VL-KnG correctly answers questions about persistent scene attributes (jacket colour, sidewalk busyness) by retrieving the relevant KG entry. It struggles when the relevant observation is brief or not prominently detected. The third example (\textcolor[rgb]{0.9,0.5,0}{\textbf{orange}}: inductive bias) highlights a case where the ground-truth label is subjective: for \textit{``Which direction did you turn after leaving the building?''}, frames 111--112 (top pair) support the VL-KnG prediction (turned right along the hallway wall), while frames 116--117 (bottom pair) support the GT annotation (turned left off the sidewalk).

\begin{figure}[h]
    \centering
    \includegraphics[width=0.96\textwidth]{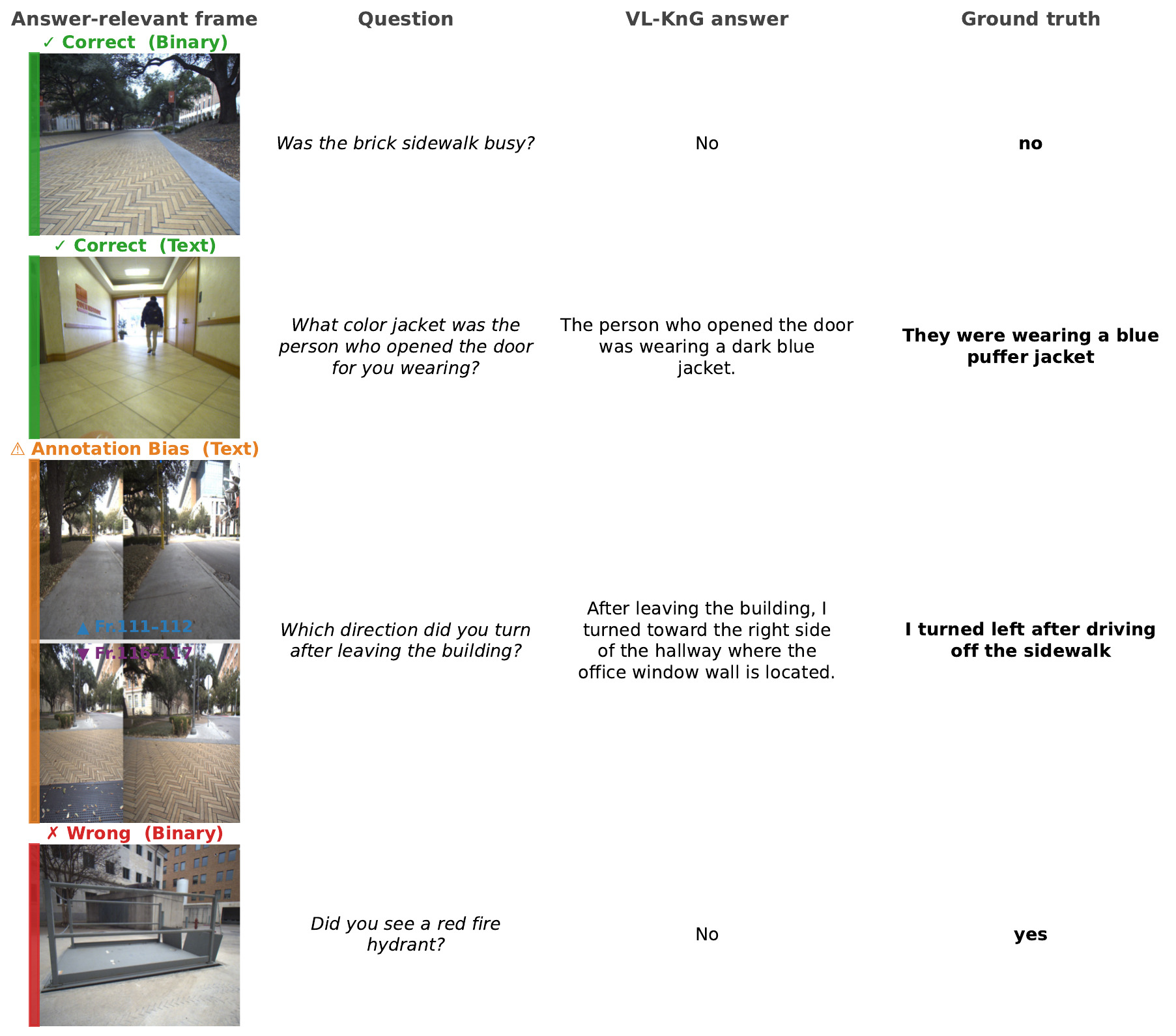}
    \caption{NaVQA qualitative examples. \textcolor[rgb]{0,0.5,0}{\textbf{Green}}: correct. \textcolor{red}{\textbf{Red}}: wrong. \textcolor[rgb]{0.9,0.5,0}{\textbf{Orange}}: inductive bias --- overlapping frame pairs show that both the VL-KnG prediction (Fr.111--112, blue) and the GT annotation (Fr.116--117, purple) are valid egocentric descriptions of the same trajectory.}
    \label{fig:navqa_qual}
\end{figure}

\clearpage
\section{Prompt Templates}
\label{ssec:prompts_supp}

VL-KnG uses a four-step prompt pipeline for knowledge graph construction and question answering. The complete prompt guide used across all experiments is reproduced below.

\begin{promptbox}[Prompt Guide: VL-KnG (Vision-Language Knowledge Graph Navigation)]
\small
\textbf{Role:} Expert at object tracking and ID consistency across video chunks.\\
\textbf{Input:} Frames (images) in chunks.\\
\textbf{Goal:} (1)~Detect objects and spatial relationships per chunk with stable IDs; (2)~Resolve IDs across chunks.

\smallskip\noindent\rule{\linewidth}{0.4pt}\smallskip

\textcolor{promptblue}{\textbf{\normalsize Step 1: Chunk Object Detection}}\\[2pt]
\textbf{Task:} Detect objects and spatial relationships across given frames. Output structured YAML.

\textcolor{promptteal}{\textbf{Critical rules:}}
\begin{itemize}
  \item \pcritical{Unique IDs:} One ID per physical object across \emph{all} frames. Format: \pcode{<type>\_<num>}. IDs increment globally.
  \item Same object in different frames $\Rightarrow$ same ID. Different objects $\Rightarrow$ different IDs.
  \item \pcritical{Spatial rels:} Extract \emph{all} spatial relationships between objects \emph{within each frame}.
  \item \pcritical{Text:} ID pattern \pcode{text\_<num>}; store exact characters in \pcode{description.content}.
\end{itemize}

\textcolor{promptteal}{\textbf{Human-centric attributes:}}
\begin{itemize}
  \item Focus on search/navigation detail: shelf contents, brands, labels, landmarks.
  \item Per object: \pcode{category}, \pcode{subcategory}, \pcode{affordance}, \pcode{area/zone}.
\end{itemize}

\textcolor{promptteal}{\textbf{Spatial relation types (allowed):}}\\
{\scriptsize\texttt{on, on\_top\_of, under, next\_to, between, in\_front\_of, behind, near, far\_from, touching, separate\_from, left\_of, right\_of, above, below, inside, outside, surrounding, adjacent\_to, against}}

\textcolor{promptteal}{\textbf{Output:}} YAML with \pcode{objects:} (id, name, description, frames with bbox) and \pcode{spatial\_relationships:} per frame.
\end{promptbox}

\begin{promptbox}[Prompt Guide (continued)]
\small
\textcolor{promptblue}{\textbf{\normalsize Step 2: Cross-Chunk ID Resolution}}\\[2pt]
\textbf{Task:} Align local chunk IDs with existing global objects.

\textcolor{promptteal}{\textbf{Matching rules:}}
\begin{itemize}
  \item Same type + similar description $\Rightarrow$ \pcritical{reuse} global ID.
  \item Clearly new object $\Rightarrow$ new unique ID.
  \item When in doubt, \pcritical{prefer reusing} an existing ID.
\end{itemize}
\textbf{Output:} Corrected YAML only. No explanations.

\medskip
\textcolor{promptblue}{\textbf{\normalsize Step 3: Question Adaptation}}\\[2pt]
\textbf{Task:} Reformulate the question using \emph{exact} object names and relationship types from the graph.

\begin{itemize}
  \item Keep \texttt{\$\$...\$\$} markers verbatim; \pcritical{do not} remove or alter.
  \item Replace vague terms with closest available object/relation.
  \item Output one concise question---no meta-explanations.
\end{itemize}

\medskip
\textcolor{promptblue}{\textbf{\normalsize Step 4: QA \& Frame Ranking}}\\[2pt]
\textbf{Task:} Given graph context, (A)~choose one candidate and rank frames, or (B)~only rank frames.

\begin{itemize}
  \item \textbf{With candidates:} Choose \emph{exactly one}. Output: \texttt{ANSWER: <candidate>} and \texttt{RANKED\_FRAMES: <indices>}.
  \item \textbf{Without candidates:} Output: \texttt{RANKED\_FRAMES: <indices>}. Minimal reasoning.
\end{itemize}
\end{promptbox}

\begin{promptbox}[Summary: Inputs \& Outputs]
\small
\begin{itemize}
  \item \textbf{Step 1:} Frames $\to$ YAML (objects + spatial relations)
  \item \textbf{Step 2:} KG YAML + chunk YAML $\to$ corrected YAML (IDs aligned)
  \item \textbf{Step 3:} Question + objects (\texttt{\$\$}) + relations $\to$ reformulated query
  \item \textbf{Step 4:} Query + graph context $\to$ \texttt{ANSWER} + \texttt{RANKED\_FRAMES}
\end{itemize}
\end{promptbox}

\endgroup

\end{document}